\documentclass{article} 
\usepackage{iclr2027_conference,times}

\usepackage{amsmath,amsfonts,bm}

\def\eqref#1{equation~\ref{#1}}

\def\1{\bm{1}}

\DeclareMathAlphabet{\mathsfit}{\encodingdefault}{\sfdefault}{m}{sl}
\SetMathAlphabet{\mathsfit}{bold}{\encodingdefault}{\sfdefault}{bx}{n}

\usepackage{hyperref}
\usepackage{url}
\usepackage{xspace}
\usepackage{amsmath}
\usepackage{algorithm}
\usepackage{algpseudocode}
\usepackage{enumitem}
\usepackage{booktabs}
\usepackage{graphicx}
\usepackage[normalem]{ulem}
\usepackage{graphicx}
\usepackage{capt-of}
\usepackage{wrapfig}

\newenvironment{tight_itemize}{\begin{itemize}[leftmargin=*] \itemsep
		1pt}{\end{itemize}}

\def\ourtask{\textsc{Video-to-4DAVScene}\xspace}

\def\ourdataset{\textsc{SonoScene4D}\xspace}
\def\ourframework{\textsc{SonoWorld4D}\xspace}

\title{\textsc{SonoWorld4D}: From One Video to a 4D Audio-Visual Scene via Diffusion Synchronization}

\author{%
  \begin{tabular}{@{}l@{}}
    Derong Jin\thanks{Equal contribution} \quad
    Xiyi Chen\footnotemark[1] \quad
    Ming C. Lin \quad
    Ruohan Gao \\
    \textnormal{University of Maryland, College Park}
  \end{tabular}%
}

\iclrfinalcopy 
\begin{document}

\maketitle

\vspace{-1.5em}
{\centering
\url{https://humathe.github.io/sonoworld4d}\par}
\vspace{1em}

\begin{abstract}
Recent progress in visual scene generation has extended beyond static 3D scenes toward the more challenging problem of generating dynamic 4D scenes from minimal input. However, sound, despite being intrinsically coupled with dynamic objects and events, remains largely unmodeled in prior work. We introduce \textsc{Video-to-4DAVScene}, a new task that turns a video into an explorable, time-varying 4D audio-visual scene, and present the \ourframework framework for this task. Given a monocular video, our method generates a 360$^\circ$ static background, reconstructs explicit motion trajectories of dynamic, sound-emitting foreground objects, and synthesizes a dynamic sound field consistent across multiple sources through dual-space diffusion synchronization. The resulting 4D scene supports real-time novel-view visual and spatial audio rendering. Experiments on real-world dynamic scenes and a user
study demonstrate that \ourframework generates
4D audio-visual scenes with perceptually realistic,
physically grounded spatial audio consistent with
scene geometry and visual content.
\\ 
\end{abstract}
\begin{figure}[H]
    \centering
    \vspace{-10pt}
    \includegraphics[width=0.99\linewidth]{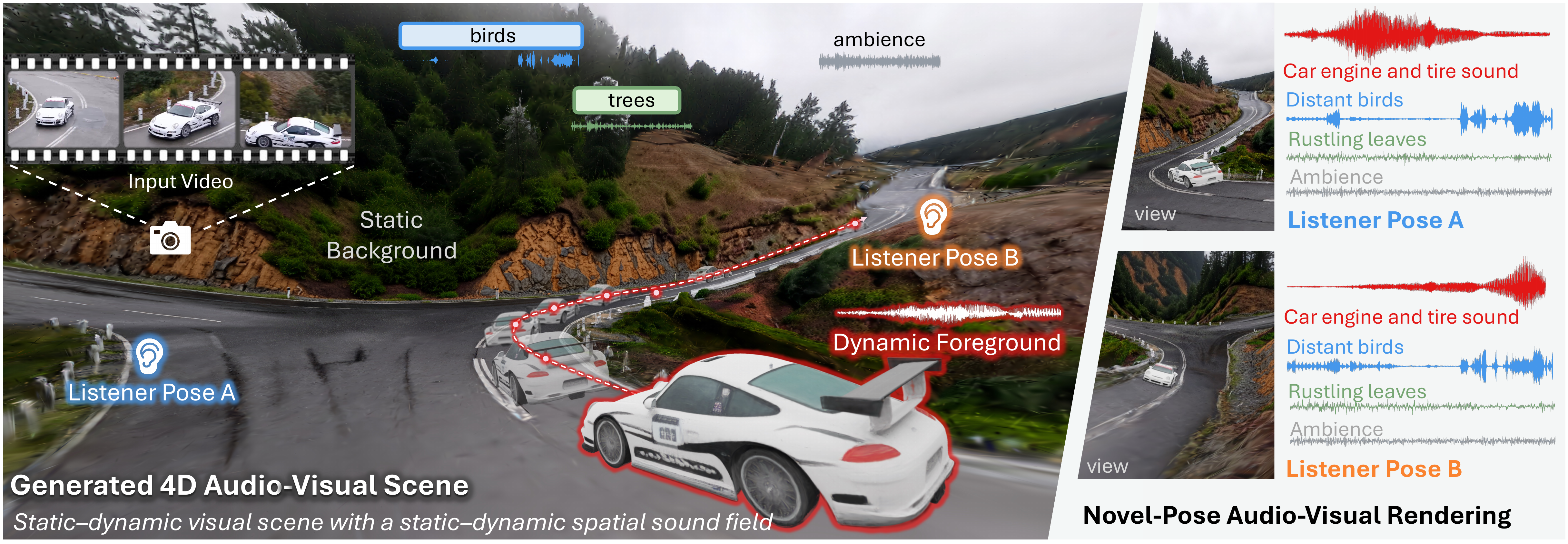}
    \vspace{-6pt}
    \caption{From a single monocular  video, \ourframework generates an explorable 4D audio-visual scene, where you can navigate to novel viewpoints and follow arbitrary listener trajectories while hearing spatial audio synchronized with dynamic events and grounded in moving sound sources (e.g., a racing car), static sources (e.g., birds and rustling trees), and background ambience.
    }
    \label{fig:teaser}
\end{figure}

\section{Introduction}\label{sec:intro}

Advances in 3D scene reconstruction and generation~\citep{nerf,3dgs,wonderworld,viewcrafter}, and more recently in dynamic 4D scene generation~\citep{wu2024cat4d, ren2025gen3c,lei2024mosca}, make it possible to transform monocular videos into explorable, time-varying worlds with moving objects and temporally coherent geometry. Yet these worlds remain incomplete without sound, which is intrinsically coupled with dynamic objects and events: 
engine and wheel noise grow louder as a car approaches, sweeping past and fading into the distance; a dancer's tap shoes strike the stage in sharp, rhythmic beats synchronized with each step; and ambient sounds such as wind and crowds continuously convey the structure and scale of the environment.

Despite its importance, limited prior work has modeled spatial audio in 3D visual scenes.
Existing approaches either focus on static scenes without time-varying geometry or moving sound sources~\citep{sonoworld,yoshida2026scene2sound}, or spatialize audio for dynamic videos under constrained camera and interaction settings~\citep{xie2025sonic4d}. Generating fully explorable 4D audio-visual scenes therefore remains an open challenge, requiring joint modeling of dynamic geometry, moving sound sources, and spatial audio. 

We introduce 
\ourtask, a new problem that aims to turn a monocular video into an explorable \emph{4D audio-visual scene}: 
an interactive, time-varying world-space scene with dynamic sound sources that supports synchronized rendering of both visual content and spatial audio from novel viewpoints and listener trajectories. 
The generated scene should contain a spatially grounded sound field: ambient sound should be tied to the static environment, dynamic sounds should follow moving visual objects, and the rendered audio should remain consistent as the virtual listener moves through the reconstructed 4D audio-visual world. 

Solving this problem introduces several new challenges. \emph{First}, unlike recent 4D visual generation methods that primarily optimize visual fidelity, 4D audio-visual scene generation requires explicit scene geometry, camera motion, dynamic object trajectories, and static-dynamic structure to place and render sound sources in a physically meaningful coordinate system. \emph{Second}, dynamic scenes contain heterogeneous acoustic components: ambient sound should align with the background, localized static sources should remain fixed in world space, and dynamic sources should follow  foreground objects while obeying distance attenuation, occlusion, and motion-dependent effects. \emph{Third}, the system must maintain consistency across coordinate spaces: input audio is observed in camera space, object sounds are represented in object-centric canonical space, and the output must be rendered for arbitrary listener trajectories in world space. These spaces must be synchronized to keep sound temporally aligned with the input video and geometrically consistent with the evolving scene.

To address these challenges, we propose \ourframework, a world-space framework for 4D audio-visual scene generation. Given an input video, we reconstruct a \emph{decomposed 4D visual scene representation}, comprising a $360^{\circ}$ static background and dynamic foreground objects with world-space trajectories, providing geometry and motion for spatial audio rendering while supporting interactive visual exploration.
Building on this representation, we introduce a \emph{diffusion-based synchronization} mechanism that jointly synthesizes a global camera-space audio signal and per-object canonical audio signals to generate a dynamic spatial sound field.  
Specifically, it repeatedly projects object-space sounds into camera space through acoustic rendering and maps the global mixture back to object space via inverse rendering, enforcing cross-space consistency throughout the diffusion process. This synchronization keeps the generated audio aligned with the input video while preserving persistent, re-renderable object sounds that follow dynamic sources under novel listener trajectories.

We evaluate \ourframework on real-world dynamic scenes with moving foreground sound sources exhibiting rigid-body, human, and general deformable motion. We also establish objective evaluation protocols for \ourtask to assess the spatial accuracy and semantic consistency of generated audio. Experiments show that our framework generates 4D audio-visual scenes with spatial audio that better aligns with scene geometry, object motion, and listener viewpoint than baseline methods. 
Ablation studies and a user study further confirm the effectiveness of diffusion synchronization in generating coherent spatial audio with multiple sound sources.

\vspace{-6pt}
\section{Related Work}\label{sec:related}
\vspace{-6pt}

\vspace{0.05in}

\noindent \textbf{4D Visual Scene Generation.} 
Recent 4D scene generation methods can be broadly grouped into video diffusion, dynamic Gaussian reconstruction, and dynamic foreground reconstruction. \emph{Video diffusion} based methods ~\citep{ren2025gen3c, xie2026lavr, Yu_2025_ICCV, lin2026vista4d, yang2026neoverse, bai2025recammaster} synthesize photorealistic videos under controlled camera trajectories, but are trajectory-specific and often lack explicit geometry grounding. 
\emph{Dynamic Gaussian reconstruction} methods ~\citep{lei2024mosca, som2024, priorenhancedgaussian} reconstruct geometry-grounded dynamic scenes from monocular videos and support real-time rendering,  
but 
are largely limited to visible regions
and cannot reliably recover unseen object surfaces or a complete panoramic background. 
\emph{Dynamic foreground reconstruction} methods further recover animatable 4D assets from monocular inputs, including category-agnostic deformable or articulated objects~\citep{pad3r, ActionMesh2026, chen2026motion3to4, chen2025v2m4, litman2026lift4d} 
and human-specific avatars~\citep{qiu2025lhmpp, hu2024gaussianavatar, moon2024exavatar}.
Our visual pipeline builds on these advances to construct a 4D visual scene representation, combining dynamic foreground objects with a static background to enable generation of a dynamic spatial sound field.

\noindent\textbf{Spatial Audio Generation.} Audio generation methods synthesize realistic monaural audio  from text 
or video~\citep{mmaudio,shan2025hunyuanvideofoleymultimodaldiffusionrepresentation,audiox,multifoley},
but typically lack spatial cues. Complementary approaches spatialize existing audio:  
RIR-based rendering and acoustic-field methods model propagation effects
~\citep{soundspace,soundspace2,naf,diffrir,avdar,avr,liu2025haae}, while visually guided methods 
use scene geometry, object locations, or motion to predict binaural or spatial audio from mono input~\citep{gao2019visual-sound,garg2023visually,garg2021geometry,morgado2018self,li2018scene,av-nerf}. 
More recent methods generate ambisonics from videos~\citep{visage,liu2025omniaudiogeneratingspatialaudio, omnidream}, spatialize generated audio for 4D scene exploration~\citep{xie2025sonic4d}, or construct  static 3D audio-visual scenes from single images~\citep{sonoworld,yoshida2026scene2sound,chen2026audible}. 
In contrast, \ourframework generates explorable dynamic 4D audio-visual scenes with moving sound sources, synchronizing spatial audio with object motion while modeling listener-dependent acoustic effects.

\noindent \textbf{Diffusion Synchronization.} 
Diffusion synchronization couples multiple diffusion trajectories at inference time to satisfy shared constraints. 
Primarily explored for image generation, it  
enables coherent large images and region-controlled layouts~\citep{multidiffusion}, enforces consistency across overlapping generations~\citep{syncdiffusion}, and models 
correlations between coupled trajectories~\citep{syncsde}. 
Recent work extends these ideas 
to synchronize image and audio diffusion on a shared canvas~\citep{chen2024images} and extend long-form audio separation beyond the training length~\citep{samaudio}. 
We also apply diffusion synchronization to audio, but target spatial audio rendering. Specifically, we synchronize camera-space and object-space audio trajectories through forward and inverse acoustic rendering, allowing generated source sounds to remain consistent with the camera-space mixture while being re-renderable under source motion and novel listener trajectories. To our knowledge, this is the first use of diffusion synchronization for spatial audio generation with an explicit rendering component.

\section{The \textsc{Video-to-4DAVScene} Task}
\label{sec:task}

We define \ourtask as the problem of generating an explorable
4D audio-visual scene from a single monocular video. Given an input video $V=\{I_t\}_{t=1}^{T},$
where $I_t$ denotes the RGB frame at time $t$, the problem is to generate: 
\begin{equation}
    V \mapsto \{\mathcal{S}, \mathcal{A}\},
\end{equation}
where $\mathcal{S}$ denotes a time-varying 4D visual scene with moving objects and $\mathcal{A}$
denotes a dynamic spatial sound field associated with the scene. The output
should be explorable: for any target camera/listener trajectory, it should
support synchronized novel-view visual rendering and listener-dependent spatial
audio rendering.

Formally, given a target trajectory $p_t = [R_t,\mathbf{x}_t] \in SE(3)$, 
where $R_t$ and $\mathbf{x}_t$ denote the listener orientation and position at
time $t$, the generated scene should render an audio-visual sequence
\begin{equation}
   \{\mathcal{S}(p_t), \mathcal{A}(p_t)\} = \left\{
    \hat{I}^{\,p}_t,\,
    \hat{a}^{\,p}_t
    \right\}.
\end{equation}
Here, $\hat{I}^{\,p}_t$ is the visual observation rendered from pose $p_t$, and
$\hat{a}^{\,p}_t$ is the synchronized spatial audio perceived by the listener at the same
pose. 
The visual scene $\mathcal{S}$ contains both the static environment and
time-varying foreground content. The spatial sound field $\mathcal{A}$ should
be geometrically and temporally grounded in $\mathcal{S}$: ambient sound and static localized sound sources should
remain fixed in static world space, and dynamic sound sources should move consistently
with their corresponding foreground objects. Thus, the generated audio should
not only match the semantic and temporal content of the input video, but also
vary coherently with listener position, listener orientation, source motion,
distance attenuation, and visibility.

\noindent\textbf{The \ourdataset Dataset.} 
To support quantitative evaluation of this new task,
we collect \ourdataset, an evaluation dataset of 54 clips of dynamic scenarios with moving sound sources such as cars and humans. Specifically, each scene contains an 8-second 360$^\circ$ video captured at 30 FPS using an Insta360 X5 camera, paired with synchronized first-order ambisonic (FOA) audio recorded at 48 kHz using a R{\O}DE NT-SF1 microphone placed at a novel listening pose. 
The FOA microphone is placed at a listening pose different from the camera pose, enabling evaluation of whether a method can render spatial audio at a novel listener location rather than merely reproducing camera-view audio. Further details about \ourdataset are provided in
Appendix (Sec.~\ref{app:dataset_details}).

\noindent\textbf{Evaluation Metrics.} 
We evaluate spatial audio generation in terms of spatial accuracy and semantic consistency.
For \emph{spatial accuracy}, we report correlation coefficient (CC) and area under the curve (AUC) between generated and reference spatial energy maps~\citep{visage}, together with direction-of-arrival (DoA) errors, including absolute azimuth error $\Delta_{\mathrm{abs}}\theta$, absolute elevation error $\Delta_{\mathrm{abs}}\phi$, and angular error $\Delta_{\mathrm{Angular}}$~\citep{liu2025omniaudiogeneratingspatialaudio}.
For \emph{semantic consistency}, we compute CLAP similarity between the rendered audio and manually annotated object descriptions.
Spatial metrics are evaluated at 5\,Hz and averaged over time, while CLAP is computed at the clip level.

\begin{figure*}[t!]
\vspace*{-0.5em}
  \centering
    \includegraphics[width=0.998\linewidth]{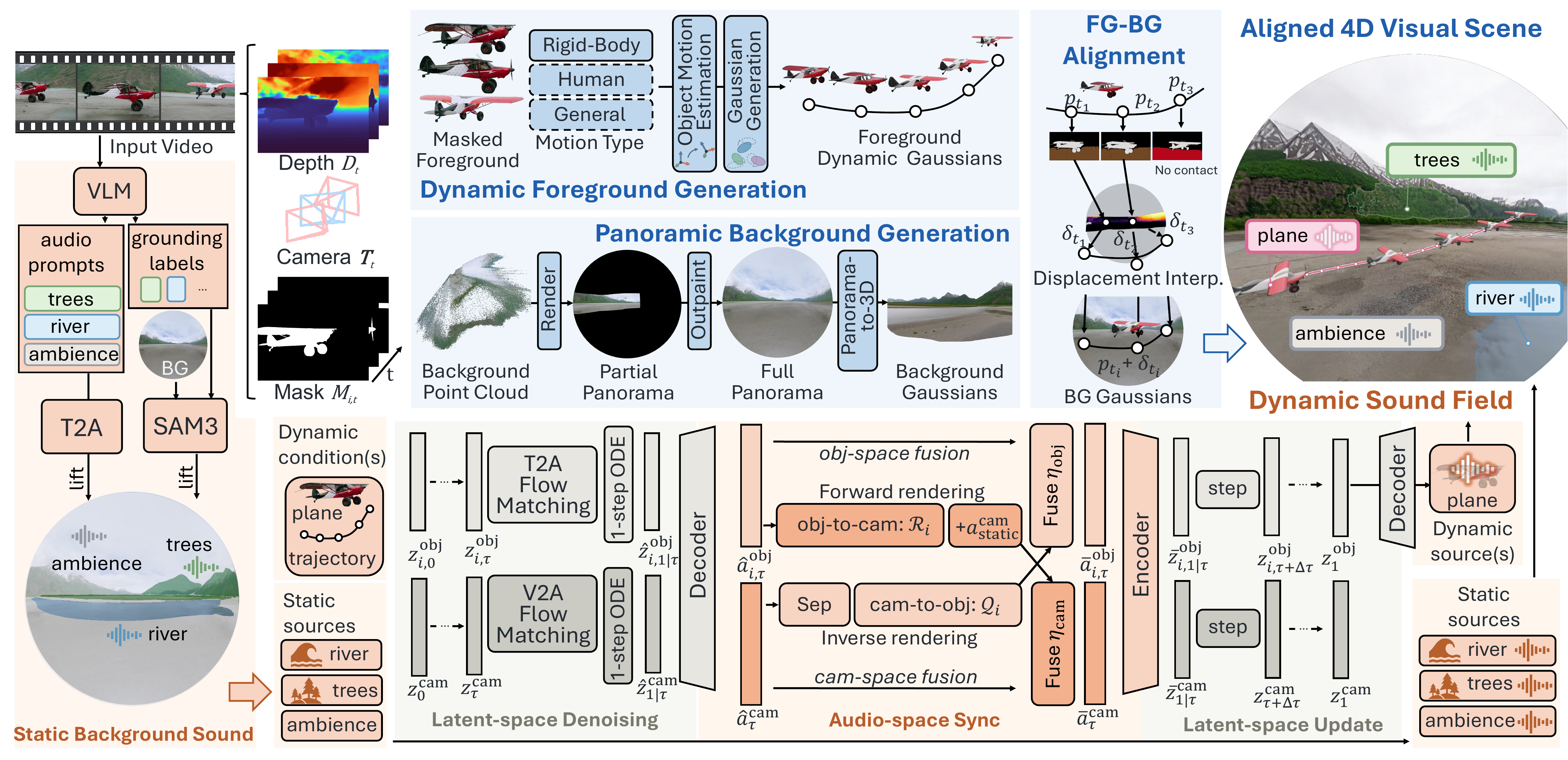}
    \vspace*{-2em}
  \caption{
  \textbf{Overview of \ourframework.} Given a monocular video, we extract depth maps, camera poses, foreground masks, and VLM-derived audio prompts. For visual scene generation (Sec.~\ref{sec:visual_decomposition}), the framework reconstructs per-object dynamic Gaussians, lifts an outpainted $360^\circ$ panorama into background Gaussians, and aligns them into a common world frame via scale correction and local displacement correction. For dynamic sound field generation (Sec.~\ref{sec:audio_field}), static sources are synthesized via text-to-audio and anchored to the background. Dynamic sources are generated via dual-space diffusion synchronization: at each sampling step, forward rendering ($\mathcal{R}_i$) maps object-space predictions to camera space, while source separation and inverse rendering ($\mathcal{Q}_i$) map camera-space mixtures back to individual object spaces. The resulting scene enables real-time novel-pose visual and spatial audio rendering (Sec.~\ref{sec:inference}).
  }
  \vspace*{-0.75em}
  \label{fig:pipeline}
\end{figure*}

\vspace{-0.1in}
\section{Approach}
\label{sec:approach}

\subsection{Overview}
Given a monocular video $V=\{I_t\}_{t=1}^{T}$, \ourframework generates an explorable 4D audio-visual scene $\{\mathcal{S}, \mathcal{A}\}$. 
We distinguish three coordinate
systems: world space is the shared, fixed reference frame
for the scene; camera space is the camera's coordinate
system at each frame; and object space is the local
coordinate system attached to each foreground object.
To support object-space audio modeling, we decompose dynamic
foreground objects from the static background and explicitly
model each object's trajectory, allowing its sound to follow
its motion even when multiple objects move independently.
All static objects are treated as part of the background.
We represent the visual scene as
$\mathcal{S}=(\mathcal{G}_{\mathrm{bg}}^{\mathrm{world}},
\{\mathcal{G}_{i}^{\mathrm{world}}\}_{i=1}^{N})$,
where $\mathcal{G}_{\mathrm{bg}}^{\mathrm{world}}$ denotes
the static background and $\mathcal{G}_{i}^{\mathrm{world}}$
denotes the $i$-th dynamic foreground object, both represented
by Gaussians that support real-time rendering.

We represent the spatial sound field as:
\begin{equation}
\mathcal{A}
=
\left(
a^{\mathrm{amb}},
\{a_j^{\mathrm{bg}}\}_{j=1}^{M},
\{a_i^{\mathrm{obj}}\}_{i=1}^{N}
\right),
\end{equation}
where $a^{\mathrm{amb}}$ denotes diffuse ambient audio,
$a_j^{\mathrm{bg}}$ denotes audio from the $j$-th static background
source, and $a_i^{\mathrm{obj}}$ denotes object-space audio from the
$i$-th dynamic source.
The reconstructed audio-visual scene provides the source locations, trajectories,
and geometry required to render these signals from arbitrary listener
poses.

Fig.~\ref{fig:pipeline} illustrates our framework. We first reconstruct a decomposed 4D visual scene in world space (Sec.~\ref{sec:visual_decomposition}): the static component provides geometry and context for ambient sound, while the dynamic component provides object-centric trajectories for moving sound sources. We then reconstruct a dynamic spatial sound field (Sec.~\ref{sec:audio_field}) by jointly generating object-space and camera-space audio through diffusion synchronization, 
allowing source sounds to remain aligned with object motion and consistent with the rendered camera mixture. During free-viewpoint rendering (Sec.~\ref{sec:inference}), a target listener trajectory is used to render novel-view visuals and spatialize the generated object sounds into synchronized audio coherently blended with the background sound. 

\vspace{-0.15in}
\subsection{4D Visual Scene 
Generation via Foreground-Background Decomposition}\label{sec:visual_decomposition}
\vspace{-0.05in}

We first preprocess the input video to estimate scene
geometry, camera poses, and foreground masks.
Conditioned on the observed background regions and visible
foreground pixels, we then generate panoramic background
Gaussians and dynamic foreground Gaussians, respectively.
Foreground generation supports three motion types:
rigid-body, human, and general deformable motion,
using a shared formulation for object motion estimation
and deformable Gaussian generation.
Finally, we align the generated foreground objects and background
in a common world frame. The resulting 4D scene provides
the geometry and object trajectories used for spatial audio
generation and rendering.

\noindent\textbf{Video Preprocessing.} 
Given an input video $\{I_t\}_{t=1}^{T}$ with $T$ frames, we use a
geometry foundation model~\citep{wang2026vggtomega} to estimate depth
maps $\{D_t\}_{t=1}^{T}$ and camera poses $\{\mathbf{T}_t\}_{t=1}^{T}$,
and use a segmentation model~\citep{carion2026sam3segmentconcepts}
to obtain foreground masks $\{M_{i,t}\}_{i=1,t=1}^{N\quad\!T}$,
where $N$ is the number of foreground objects. 
We further use a vision-language model (VLM)~\citep{team2026kimi}
to infer acoustic context from the video. It produces a video-level
description $d_{\mathrm{vid}}$ capturing the overall acoustic scene,
including ambient and background sounds, and an object-level description
$d_i$ for each sound-producing foreground object $i$, capturing its
sound semantics and temporal activity. The former conditions camera-space
audio generation, while the latter conditions the corresponding
object-space source generation in Sec.~\ref{sec:audio_field}.

\noindent\textbf{Panoramic Background Generation.} 
We generate the background from observed background
regions in the input video. To consolidate these observations into
a single panorama for completion, we select the frame with the largest visible background region
as an anchor frame and take its coordinate system as the world frame. We then unproject background pixels from the depth maps across all
frames and rasterize the aggregated points in the world frame to
obtain a partial background panorama
$I_{\mathrm{pano}}^{\mathrm{partial}}$.
Next, we apply single-camera calibration~\citep{geocalib} to align
the panorama with the gravity direction and use a panorama diffusion
outpainting model~\citep{nomadoor2026flux2panorama,flux2} to generate
a complete $360^\circ$ background panorama
$I_{\mathrm{pano}}^{\mathrm{full}}$.
Finally, a panorama-to-3D model~\citep{hyworld2026, worldlabs_marble} lifts it into background Gaussians $\mathcal{G}_{\mathrm{bg}}^{\mathrm{world}}$.

\noindent\textbf{Dynamic Foreground Generation.} We generate a dynamic Gaussian representation for each foreground object conditioned on its visible pixels in the input video. Dynamic object modeling remains an active area of research~\citep{ActionMesh2026,pad3r,chen2026motion3to4,helix4d2026},
with visual backbones offering different capabilities across motion types. We therefore adopt a unified formulation with motion-type-specific instantiations for three categories: 
\textit{rigid-body motion}, which preserves object shape; \textit{human motion}, which includes body articulation; and \textit{general deformable motion}, which allows flexible shape changes without assuming a human body model. 
Across all three types, we model global motion through per-frame scale, rotation, and translation, while treating local deformation differently. 

Generation comprises two shared components: \emph{i) Object Motion Estimation:} 
For each foreground object $i$ and frame $t$, we estimate a
similarity transformation
$\mathbf{H}_{i,t}^{\mathrm{obj}\to\mathrm{cam}}
= (\alpha_{i,t}, \mathbf{R}_{i,t}, \mathbf{p}_{i,t})
\in \mathrm{Sim}(3)$,
where $\alpha_{i,t}$, $\mathbf{R}_{i,t}$, and $\mathbf{p}_{i,t}$
denote scale, rotation, and translation, respectively.
This transformation maps the object's locally deformed geometry from object space into camera space at frame $t$. These parameters describe its scale, orientation, and position, while local shape changes are modeled by the second component, described next. 

\emph{ii) Deformable Gaussian Generation:}
Conditioned on visible foreground pixels of object $i$, we generate
its canonical Gaussians $\mathcal{G}_i^{\mathrm{obj}}$ and a time-dependent deformation field $\Phi_{i,t}$. This field maps $\mathcal{G}_i^{\mathrm{obj}}$ to their deformed
configuration in object space at frame $t$, capturing
local articulation or shape changes, and reduces to
the identity for rigid objects. We apply $\Phi_{i,t}$,
followed by the object-to-camera transformation
$\mathbf{H}_{i,t}^{\mathrm{obj}\to\mathrm{cam}}$
and the camera-to-world transformation
$\mathbf{T}_t^{\mathrm{c}\to\mathrm{w}}$, to obtain
dynamic foreground Gaussians
$\mathcal{G}_{i,t}^{\mathrm{world}}$.
The full formulation and motion-type-specific
instantiations are provided in
Appendix (Sec.~\ref{app:motion_instantiations}).
The resulting foreground Gaussians and object
trajectories are then aligned with the
panoramic background.

\noindent\textbf{Foreground-Background Alignment.} 
Since the foreground and background are generated
independently, their scales and local geometry may
be inconsistent. We align the independently generated foreground
and background through global scale alignment and
local displacement correction. We first estimate
a background scale $s$ in closed form by matching
the generated background to the video-derived point
cloud over their common visible region.
For each foreground object, we then estimate
displacements $\delta_{i,t}$ from correspondences
between its local contact neighborhood and the
scale-aligned background. For frames without reliable
contact, we interpolate the displacements from
supported frames, retaining the nearest displacement
outside their temporal range.
We apply these corrections to the object's world-space
positions, $\mathbf{p}_{i,t}^{\mathrm{world}}
+\delta_{i,t}$, and its Gaussians to obtain an aligned
4D visual scene. 
Details are provided in Appendix (Sec.~\ref{app:fg_bg_alignment}).

\vspace{-0.15in}
\subsection{Dynamic Spatial Sound Field Generation via Diffusion Synchronization}
\label{sec:audio_field}
\vspace{-0.05in}

The reconstructed 4D visual scene provides the geometry and trajectories of sound
sources, while the video provides their semantic and temporal context.
Our goal is to generate persistent source signals that can be spatially
re-rendered along arbitrary listener trajectories while remaining consistent
with the acoustic events observed in the input video.

To this end, we represent audio in two complementary spaces.
We denote by $a^{\mathrm{cam}}$ a video-conditioned camera-space mixture,
which captures the overall acoustic content observed by the input camera,
and by $a_i^{\mathrm{obj}}$ the object-space signal of dynamic source $i\in\mathcal{I}_{\mathrm{dyn}}$,
which can be independently re-rendered from novel listener poses.
Generating these signals independently, however, does not guarantee that
the object tracks explain the camera-space mixture.
We therefore couple the two representations through the acoustic renderer
induced by the reconstructed 4D scene.

\noindent\textbf{Static Background Sound Generation.} We generate background audio using the VLM-derived descriptions and grounding labels obtained during video preprocessing. For each static source, a text-to-audio model~\citep{shan2025hunyuanvideofoleymultimodaldiffusionrepresentation} synthesizes a source signal conditioned on its textual description. To spatially ground the signal, we use its grounding label to prompt SAM3~\citep{carion2026sam3segmentconcepts} for segmentation in the background panorama and associate the resulting region with the reconstructed background geometry, obtaining a point cloud as a fixed spatial anchor for the source. Global ambience is maintained as a separate, non-spatialized background track. Together, the rendered static-source signals and ambience provide the background contribution $a_\text{static}^\text{cam}$ to the camera-space mixture used for diffusion synchronization (Eq.~\ref{eq:object_to_camera}).

\noindent\textbf{Bidirectional Camera-Object Consistency. } For each dynamic source $i\in\mathcal I_{\mathrm{dyn}}$, the reconstructed
scene defines an acoustic rendering operator $\mathcal R_i$ that maps its
object-space signal to its contribution along the observed camera trajectory. We focus on outdoor scenes with limited reverberation and model direct-path propagation with Doppler shifts, distance attenuation, and air absorption, omitting reflected paths.
Rendering and summing the source tracks gives a camera-space prediction:
\begin{equation}
    \tilde a^{\mathrm{cam}}
    =
    \sum_{i\in\mathcal I_{\mathrm{dyn}}}
    \mathcal R_i\!\left(a_i^{\mathrm{obj}}\right)
    + a^{\mathrm{cam}}_{\mathrm{static}},
    \label{eq:object_to_camera}
\end{equation}
where $a^{\mathrm{cam}}_{\mathrm{static}}$ denotes the
rendered static-source and ambient contributions.
We require this rendered mixture to remain consistent with the
video-conditioned prediction $a^{\mathrm{cam}}$.

Conversely, the camera-space mixture provides a video-aligned cue for each
dynamic source. We first isolate source $i$ using a query-conditioned
separator $\operatorname{Sep}_{q_i}$ and then approximately invert its acoustic
propagation:
\begin{equation}
    \tilde a_i^{\mathrm{obj}}
    =
    \mathcal {Q}_i
    \!\left(
        \operatorname{Sep}_{q_i}
        \!\left(a^{\mathrm{cam}}\right)
    \right),
    \label{eq:camera_to_object}
\end{equation}
where $q_i$ is the source-specific separation prompt and
$\operatorname{Sep}(\cdot)$ is instantiated with SAM-Audio~\citep{samaudio}.
Here, $\mathcal{Q}_i$ is the inverse renderer that maps the separated camera-space signal back to object space. Together, Eqs.~\ref{eq:object_to_camera} and \ref{eq:camera_to_object} establish bidirectional camera–object consistency, applying the object-space constraint. Static and ambient tracks contribute to the forward camera-space mixture but do not receive camera-to-object synchronization feedback. We detail the complete inverse formulation and its derivation in the Appendix (Sec.~\ref{app:forward_inverse_acoustic}).

\noindent\textbf{Dual-Space Diffusion Synchronization.} We enforce the above constraints during sampling by synchronizing one
camera-space diffusion trajectory with the object-space trajectories of all
dynamic sources. We use the flow-matching formulation of
HY-VideoFoley~\citep{shan2025hunyuanvideofoleymultimodaldiffusionrepresentation}.
At each flow step, we first predict clean audio for every branch, synchronize
the predictions in audio space through Eq.~\ref{eq:object_to_camera} and
Eq.~\ref{eq:camera_to_object}, and then map the synchronized predictions back
to the latent flow. 

\emph{Clean prediction.}
Let $\tau\in[0,1)$ denote flow time, and let $z_\tau^k$ and $v_\tau^k$
denote the latent state and predicted velocity of branch $k$, where
$k\in\{\mathrm{cam}\}
\cup
\mathcal I_{\rm dyn}$.
We extrapolate each branch to its predicted clean endpoint and decode it with the audio decoder $\operatorname{Dec}$:
\begin{equation}
    \hat z_{1|\tau}^k = z_\tau^k + (1-\tau)v_\tau^k,\qquad \hat a_\tau^k = \operatorname{Dec}\!\left(\hat z_{1|\tau}^k\right).
    \label{eq:audio_clean_prediction}
\end{equation}

\emph{Audio-space synchronization.}
We first render the object-space predictions into camera space:
\begin{equation}
    \tilde a_\tau^{\mathrm{cam}}
    =
    \sum_{i\in\mathcal I_{\mathrm{dyn}}}
    \mathcal R_i
    \left(\hat a_{i,\tau}^{\mathrm{obj}}\right)
    + a_{\mathrm{static}}^{\mathrm{cam}}.
\end{equation}
We synchronize the camera-space prediction using
\begin{equation}
    \bar a_\tau^{\mathrm{cam}}
    =
    (1-\eta_{\mathrm{cam}})
    \hat a_\tau^{\mathrm{cam}}
    +
    \eta_{\mathrm{cam}}
    \tilde a_\tau^{\mathrm{cam}},
    \qquad
    \eta_{\mathrm{cam}}\in[0,1].
    \label{eq:camera_sync}
\end{equation}

Conversely, the camera-space prediction provides an object-space hint
$\tilde a_{i,\tau}^{\mathrm{obj}}
    =
    \mathcal Q_i
    \left(
         \operatorname{Sep}_{q_i}
        \left(\hat a_\tau^{\mathrm{cam}}\right)
    \right)$.
The synchronized object-space target is then
\begin{equation}
    \bar a_{i,\tau}^{\mathrm{obj}}
    =
    \left(1-\eta_{\mathrm{obj}}\right)
    \hat a_{i,\tau}^{\mathrm{obj}}
    +
    \eta_{\mathrm{obj}}
    \tilde a_{i,\tau}^{\mathrm{obj}},
    \qquad
    \eta_{\mathrm{obj}}\in[0,1].
    \label{eq:object_sync}
\end{equation}

\emph{Synchronized flow update.}
We encode each synchronized audio target into a clean latent endpoint $\bar z_{1|\tau}^k
    =
    \operatorname{Enc}\!\left(\bar a_\tau^k\right)$, 
and replace the branch velocity by the direction toward this synchronized
endpoint:
\begin{equation}
    v_{\mathrm{sync},\tau}^k
    =
    \frac{
        \bar z_{1|\tau}^k-z_\tau^k
    }{1-\tau},
    \qquad
    \frac{\mathrm d z_\tau^k}{\mathrm d\tau}
    =
    v_{\mathrm{sync},\tau}^k.
    \label{eq:audio_sync_flow}
\end{equation}
Repeating this procedure throughout sampling couples the video-conditioned
camera trajectory with the object-conditioned source trajectories, producing
source signals that remain both video-consistent and independently
re-renderable.

\vspace*{-0.5em}
\subsection{Free-Viewpoint Rendering of Dynamic Scenes}
\label{sec:inference}
Given a target camera/listener trajectory $p_t=[R_t,\mathbf{x}_t]\in SE(3)$, we render the visual and audio streams in the same world coordinate system. For visuals, we render the static background and dynamic foreground objects at time $t$ using the reconstructed Gaussian splats, compositing the foreground objects according to their estimated trajectories $\mathbf{X}_i(t)$. For audio, we first render each localized source signal into the listener coordinate frame using the acoustic renderer, which accounts for source position, motion, attenuation, and visibility. The resulting directional source contributions are then encoded into ambisonics:
\begin{equation}
    a_L(t)
    =
    \sum_{i\in\mathcal{I}_{\text{dyn}}\cup\mathcal{I}_{\text{static}}}
    \sigma_i(t)\,
    a_i(t_i^{\mathrm{em}})
    \,
    y_L\!\left(
    R_t^\top
    \frac{\mathbf{s}_i(t)-\mathbf{x}_t}
    {\|\mathbf{s}_i(t)-\mathbf{x}_t\|}
    \right),
\end{equation}
where $\mathcal{I}_{\mathrm{dyn}}$ and $\mathcal{I}_{\mathrm{static}}$
index dynamic foreground sources and localized static background
sources, respectively, and $a_i$ denotes the corresponding source signal. The world-space source position $\mathbf{s}_i(t)$ follows the reconstructed trajectory for dynamic sources and remains fixed for static sources. Here, $t_i^{\mathrm{em}}$ is the emission time,
$\sigma_i(t)$ models distance attenuation and occlusion, and $\mathbf{y}_L$ denotes spherical harmonics up to order $L$. Finally, the ambisonics signal is decoded to binaural audio with an HRTF decoder:
\begin{equation}
    b(t)
    =
    \sum_{\ell=0}^{L}
    \sum_{m=-\ell}^{\ell}
    h_{\ell,m} \star a_{\ell,m}(t),
\end{equation}
where $h_{\ell,m}$ are the left/right HRIR filters for each
ambisonics channel. Global ambience is then mixed with $b(t)$ to produce the final
binaural output. This produces synchronized novel-view video and listener-dependent spatial audio along the target trajectory.

\begin{table*}[t]
\begin{small}
\centering
\begin{tabular}{lccccccc}
\toprule
& \multicolumn{5}{c}{Spatial Metrics}
& \multicolumn{2}{c}{Semantic Metrics}\\
\cmidrule(lr){2-6} \cmidrule(lr){7-8}
Method
& $\Delta_{\text{abs}}\theta$ $\downarrow$
& $\Delta_{\text{abs}}\varphi$ $\downarrow$
& $\Delta_{\text{Angular}}$ $\downarrow$
& CC $\uparrow$
& AUC $\uparrow$
& CLAP$_{\text{A}}$ $\uparrow$
& CLAP$_{\text{T}}$ $\uparrow$\\
\midrule
\textsc{ViSAGe}
& 74.1 & 24.5 & 75.9 & 0.189 & 0.615 & 0.078 & $-$0.059 \\
\textsc{OmniAudio}
& 80.9 & 30.0 & 83.1 & 0.124 & 0.579 & 0.234 & 0.057 \\
\textsc{SonoWorld}
 & 65.7 & 24.3 & 66.6 & 0.333 & 0.672 & 0.365 & 0.154\\
\textsc{Sonic4D}
& 54.3 & 17.2 & 55.5  & 0.446 & 0.728 & 0.223 & 0.103 \\
\midrule
Ours (T2A)
& 45.7 & \textbf{16.8} & 48.4 & 0.566 & 0.787 & 0.349 & 0.185 \\
Ours (V2A)
& 43.2 & 16.9 & 46.1 & 0.591 & 0.800 & 0.527 & 0.261 \\
\midrule
Ours
& \textbf{36.6}
& 16.9
& \textbf{40.7}
& \textbf{0.647}
& \textbf{0.835}
& \textbf{0.560}
& \textbf{0.274} \\
\bottomrule
\end{tabular}
\caption{\textbf{Quantitative comparison on \ourdataset
for video-to-4D audio-visual scene generation.}
We compare our full method against baselines and two ablated variants: Ours (T2A) and Ours (V2A), which use text-to-audio and video-to-audio generation for dynamic-source signals, respectively. Angular errors are in degrees. $\downarrow$/$\uparrow$ indicate whether lower/higher is better.
}
\label{tab:quantitative4d}
\end{small}
\end{table*}

\begin{figure}[th]
\vspace*{-0.5em}
    \centering
\includegraphics[width=0.99\linewidth]{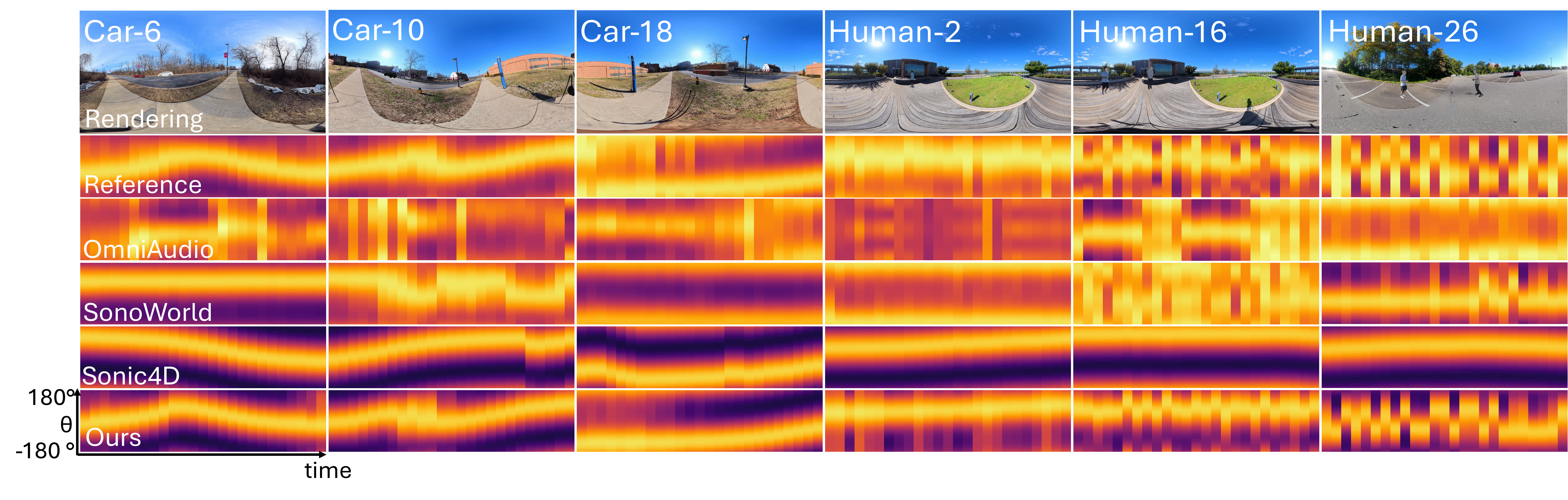}
\vspace*{-1em}
    \caption{\textbf{Time-azimuth RMS visualization of ambisonics.} Top:   mid-frame panoramas of each video rendered by our method at microphone poses in \ourdataset with faces blurred to protect privacy. 
    Remaining rows compare the reference FOA recordings with baselines and ours. Each map shows the peak-normalized directional RMS response over time and azimuth after spherical-area-weighted aggregation over elevation. Our method more closely follows the reference azimuthal trajectories and temporal structure, while the baselines exhibit directional mismatches or reduced spatiotemporal variation. 
    }
    \vspace*{-1em}
    \label{figure:rms}
\end{figure}

\section{Experiments}
\label{sec:exps}

\subsection{Experiment Settings}

We evaluate on our newly collected dataset, \ourdataset, with metrics introduced in Sec.~\ref{sec:task}, by rendering spatial audio at the microphone pose and comparing it with the ground-truth FOA recording. We compare against four prior methods that generate spatial audio from visual inputs: 

\begin{figure}[htb]
    \centering
    \includegraphics[width=0.95\linewidth]{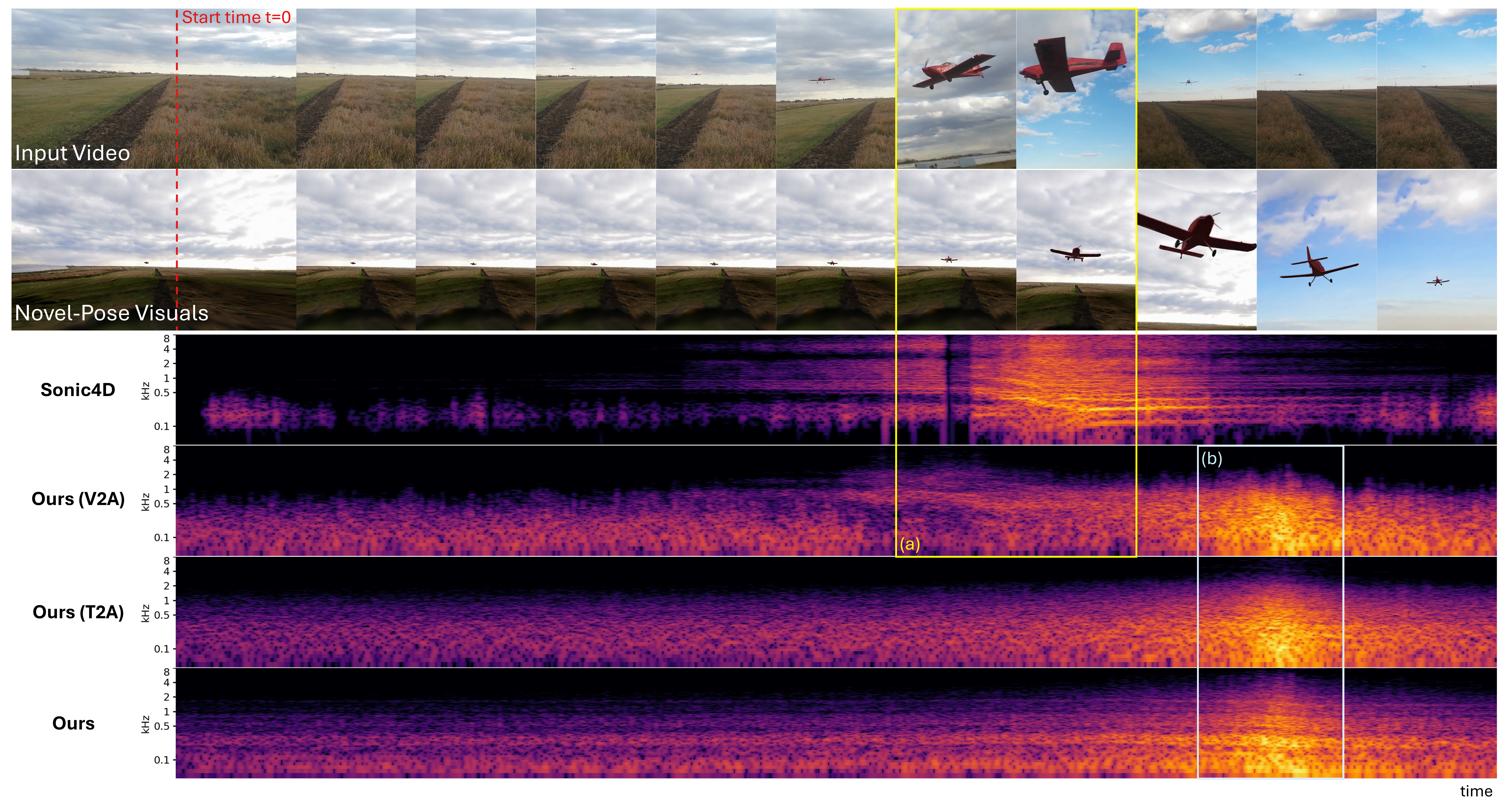}
    \vspace*{-1.5em}
    \caption{
\textbf{STFT visualization of spatial audio rendered at a novel-listener position.} The top two rows show input video frames and our novel-view renderings on a shared timeline. Highlighted intervals (a) and (b) indicate when the airplane is closest to the input camera and the novel listener, respectively. Correct spatialization should concentrate energy near (b) with a downward frequency shift due to the Doppler effect. Sonic4D retains strong energy near (a). Ours (V2A) responds near both (a) and (b). Ours (T2A) responds near (b) but lacks clear spectral structure. Our full method produces a dominant response with a visible Doppler shift around (b), correctly adapting both amplitude and frequency to the novel-listener position.
}
    \vspace*{-1em}
    \label{figure:stft}
\end{figure}

\vspace{-3pt}
\begin{tight_itemize}
    \item \textsc{ViSAGe}~\citep{visage}: 
    generates ambisonics at the input
camera pose from a field-of-view (FoV) video, without
supporting novel-view audio rendering. To evaluate it at
the target microphone pose, we provide a FoV video rendered
from our 4D visual scene at that pose.

    \vspace*{0.1em}
    \item \textsc{OmniAudio}~\citep{liu2025omniaudiogeneratingspatialaudio}: 
    generates ambisonics at the input
camera pose from a panoramic video, likewise without
supporting novel-view audio rendering. We therefore provide
a panoramic video rendered from our 4D visual scene at
the target microphone pose.

    \vspace*{0.1em}
    \item \textsc{SonoWorld}~\citep{sonoworld}:
    generates a static 3D audio-visual scene from a single image.
Since it does not model scene dynamics, we use the middle
frame of the input video and otherwise follow its original
setting for evaluation.
 
    \vspace*{0.1em}
    \item \textsc{Sonic4D}~\citep{xie2025sonic4d}:
    is the closest prior method for spatializing video-conditioned monaural audio in 4D scenes. It supports novel-view rendering but assumes a single sound source. 
We retain its visual backbone and audio pipeline,
track the dominant moving sound source, and render
at the target microphone pose. For comparison with
ground-truth FOA, we adapt its binaural
renderer using one omnidirectional and three
co-located, orthogonal figure-eight virtual microphones.
    \vspace*{-0.6em}
\end{tight_itemize}

We also compare with two variants of our method: \emph{Ours (T2A)}, which uses
text-to-audio generation for each dynamic-source track, and
\emph{Ours (V2A)}, which instead uses video-to-audio generation
conditioned on the input video.

\vspace*{-1em}
\subsection{Quantitative Results}

\textbf{Quantitative Evaluation on \ourdataset. }
Tab.~\ref{tab:quantitative4d} shows results on  \ourtask using the metrics defined in Sec.~\ref{sec:task}. 
Compared with prior video-to-audio generation baselines, including \textsc{ViSAGe}~\citep{visage} and \textsc{OmniAudio}~\citep{liu2025omniaudiogeneratingspatialaudio}, \ourframework achieves substantially better performance. 
Compared with \textsc{SonoWorld} and \textsc{Sonic4D}~\citep{xie2025sonic4d}, we also show large gains,
confirming the effectiveness of our explicit object-space audio generation and dual-space synchronization mechanisms, which improve spatial audio consistency over a trajectory-based dynamic spatialization baseline.

\begin{wrapfigure}{r}{0.49\columnwidth}
\centering
\vspace{-\intextsep}
\includegraphics[width=\linewidth]{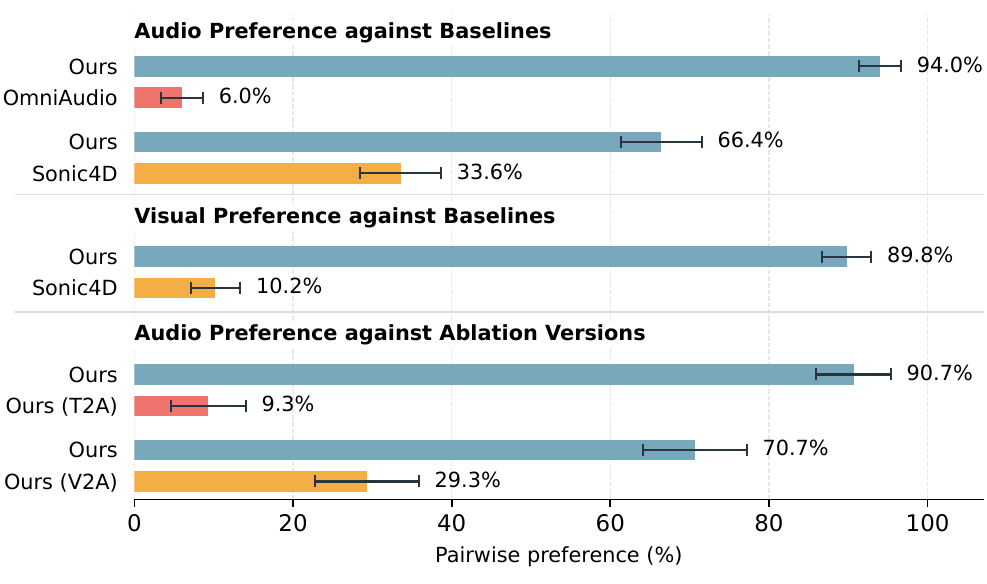}
\setlength{\abovecaptionskip}{3pt}
\vspace{-1.5em}
\caption{
\textbf{Pairwise user preferences}.
Bars show mean per-participant preference rates,
and error bars indicate 95\% confidence intervals.
}
\label{fig:user_study}
\vspace{-0.15in}
\end{wrapfigure}

\textbf{User Study. } We conduct a user study with 50 participants comparing our method against baselines and ablated variants across nine dynamic scenes from our dataset and web videos,  covering multi-source scenarios and rigid-body, human, and general deformable motion.
Participants use headphones to assess audio-visual synchronization, spatial direction and distance cues, audio quality, and consistency with a shared sound source description. 
Comparisons with \textsc{Sonic4D} also assess overall visual quality. As shown in Fig.~\ref{fig:user_study}, participants prefer our audio over \textsc{OmniAudio} and \textsc{Sonic4D} in 94.0\% and 66.4\% of comparisons, respectively, and our visuals over \textsc{Sonic4D} in 89.8\%. Our full method is also preferred over the T2A and V2A variants in 90.7\% and 70.7\% of audio comparisons, respectively. See appendix (Sec.~\ref{app:user_study}) for details.

\subsection{Qualitative Results}

\textbf{Time-Azimuth RMS Visualization.} Fig.~\ref{figure:rms} visualizes time–azimuth RMS maps of generated ambisonics for scenes in \ourdataset, alongside reference FOA recordings at the target microphone poses. Brighter regions indicate stronger directional responses. Our maps more closely follow the reference patterns across both vehicle and human scenes. In vehicle scenes, the bright bands track the changing sound direction as vehicles move through the scene; in human scenes, the maps retain intermittent responses aligned with repeated acoustic events. The baselines exhibit misplaced directional responses, overly persistent bands, or reduced temporal variation, confirming that our method better captures both the evolving sound directions and the temporal structure.

\textbf{Novel-Listener Audio Adaptation.} Fig.~\ref{figure:stft} further visualizes audio generated for a novel listener position in the airplane scene, alongside the input video and our novel-view visual renderings. Intervals (a) and (b) indicate when the airplane is closest to the input camera and the novel listener, respectively. At the novel listener position, the strongest response should occur near (b), with frequencies shifting downward as the airplane passes. \textsc{Sonic4D} retains a strong response near (a). Ours (V2A) responds near both intervals, as its generated audio already encodes Doppler shifts and loudness variations tied to the input camera position, which persist when re-rendered for the novel listener. Ours (T2A) responds near (b) but with less distinct spectral structure. Our full method produces a dominant response and a clear downward frequency shift around (b), consistent with the expected Doppler effect. These results confirm that the synchronization scheme effectively adapts both the amplitude envelope and motion-dependent frequency content to the novel listener position.

\vspace*{-1em}
\section{Conclusion}
\label{sec:conclusion}

We presented \ourtask, a new task that transforms a monocular video into an explorable, time-varying 4D audio-visual scene, and \ourframework, a training-free framework for this task. Our framework combines a 4D visual scene representation that decomposes scenes into static backgrounds and dynamic foreground objects in a unified world coordinate system, with a novel diffusion synchronization mechanism tailored for dynamic spatial sound field generation. This enables real-time interactive exploration with moving sound sources and temporally, semantically, and geometrically aligned visual and spatial audio rendering. Experiments on real-world dynamic scenes demonstrate that \ourframework outperforms a series of baselines on both objective metrics and in a user study. Future work will explicitly model object materials to support more physically accurate rendering of interaction-driven sounds in dynamic scenes.

\subsection*{Acknowledgments} 
This work was supported in part by faculty research awards from Sony and Dolby Laboratories and by the Barry Mersky and Capital One Endowed Professorships. 

\subsection*{AI use statement} 
We used AI tools to refine author-written text and assist
with code development for this work. The authors reviewed
and revised the AI-assisted text and code and take full
responsibility for the final manuscript and implementation.

\bibliography{main}

@String{Computer = "{IEEE} Computer" }

@String(IJCV  = {Int. J. Comput. Vis.})

@String(CVPR  = {IEEE Conf. Comput. Vis. Pattern Recog.})

@String(ICCV  = {Int. Conf. Comput. Vis.})

@String(ECCV  = {Eur. Conf. Comput. Vis.})

@String(NeurIPS = {Adv. Neural Inform. Process. Syst.})

@String(ICML  = {Int. Conf. Mach. Learn.})

@String(ICLR  = {Int. Conf. Learn. Represent.})

@String(BMVC  = {Brit. Mach. Vis. Conf.})

@String(TOG   = {ACM Trans. Graph.})

@String(IJCV  = {IJCV})

@String(CVPR  = {CVPR})

@String(ICCV  = {ICCV})

@String(ECCV  = {ECCV})

@String(NeurIPS = {NeurIPS})

@String(ICML  = {ICML})

@String(ICLR  = {ICLR})

@String(BMVC  =	{BMVC})

@String(TOG   = {ACM TOG})

@article{samaudio,
    title={SAM Audio: Segment Anything in Audio},
    author={Bowen Shi and Andros Tjandra and John Hoffman and Helin Wang and Yi-Chiao Wu and Luya Gao and Julius Richter and Matt Le and Apoorv Vyas and Sanyuan Chen and Christoph Feichtenhofer and Piotr Doll{\'a}r and Wei-Ning Hsu and Ann Lee},
    year={2025}
}

@inproceedings{nerf,
 title={NeRF: Representing Scenes as Neural Radiance Fields for View Synthesis},
 author={Ben Mildenhall and Pratul P. Srinivasan and Matthew Tancik and Jonathan T. Barron and Ravi Ramamoorthi and Ren Ng},
 year={2020},
 booktitle={ECCV},
}

@article{3dgs,
      author       = {Kerbl, Bernhard and Kopanas, Georgios and Leimk{\"u}hler, Thomas and Drettakis, George},
      title        = {3D Gaussian Splatting for Real-Time Radiance Field Rendering},
      journal      = {TOG},
      year         = {2023}
}

@inproceedings{wonderworld,
    title={WonderWorld: Interactive 3D Scene Generation from a Single Image},
    author={Hong-Xing Yu and Haoyi Duan and Charles Herrmann and William T. Freeman and Jiajun Wu},
    booktitle={CVPR},
    year={2025}
}

@inproceedings{sonoworld,
    title={SonoWorld: From One Image to a 3D Audio-Visual Scene},
    author={Derong Jin and Xiyi Chen and Ming Lin and Ruohan Gao},
    booktitle={CVPR},
    year={2026}
}

@article{viewcrafter,
    title={ViewCrafter: Taming Video Diffusion Models for High-fidelity Novel View Synthesis},
    author={Yu, Wangbo and Xing, Jinbo and Yuan, Li and Hu, Wenbo and Li, Xiaoyu and Huang, Zhipeng and Gao, Xiangjun and Wong, Tien-Tsin and Shan, Ying and Tian, Yonghong},
    journal={arXiv preprint arXiv:2409.02048},
    year={2024}
  }

@InProceedings{Yu_2025_ICCV,
    author    = {Yu, Mark and Hu, Wenbo and Xing, Jinbo and Shan, Ying},
    title     = {TrajectoryCrafter: Redirecting Camera Trajectory for Monocular Videos via Diffusion Models},
    booktitle = {ICCV},
    year      = {2025}
}

@inproceedings{lin2026vista4d,
    author = {Lin, {Kuan Heng} and Liu, Zhizheng and Salamanca, Pablo and Kant, Yash and Burgert, Ryan and Namekata, Koichi and Zhao, Yiwei and Zhou, Bolei and Goldblum, Micah and Debevec, Paul and Yu, Ning},
    booktitle = {CVPR},
    title = {{Vista4D}: Video Reshooting with 4D Point Clouds},
    year = {2026}
}

@inproceedings{ren2025gen3c,
    title={GEN3C: 3D-Informed World-Consistent Video Generation with Precise Camera Control},
    author={Ren, Xuanchi and Shen, Tianchang and Huang, Jiahui and Ling, Huan and 
        Lu, Yifan and Nimier-David, Merlin and M\"uller, Thomas and Keller, Alexander and 
        Fidler, Sanja and Gao, Jun},
    booktitle={CVPR},
    year={2025}
}

@article{lei2024mosca,
  title={MoSca: Dynamic Gaussian Fusion from Casual Videos via 4D Motion Scaffolds},
  author={Lei, Jiahui and Weng, Yijia and Harley, Adam and Guibas, Leonidas and Daniilidis, Kostas},
  journal={arXiv preprint arXiv:2405.17421},
  year={2024}
}

@article{wu2024cat4d,
    title={{CAT4D: Create Anything in 4D with Multi-View Video Diffusion Models}},
    author={Wu, Rundi and Gao, Ruiqi and Poole, Ben and Trevithick, Alex and Zheng, Changxi and Barron, Jonathan T. and Holynski, Aleksander},
    journal={arXiv:2411.18613},
    year={2024}
}

@article{xie2025sonic4d,
  title={Sonic4D: Spatial Audio Generation for Immersive 4D Scene Exploration},
  author={Xie, Siyi and Zhu, Hanxin and He, Tianyu and Li, Xin and Chen, Zhibo},
  journal={arXiv preprint arXiv:2506.15759},
  year={2025}
}

@misc{carion2026sam3segmentconcepts,
      title={SAM 3: Segment Anything with Concepts}, 
      author={Nicolas Carion and Laura Gustafson and Yuan-Ting Hu and Shoubhik Debnath and Ronghang Hu and Didac Suris and Chaitanya Ryali and Kalyan Vasudev Alwala and Haitham Khedr and Andrew Huang and Jie Lei and Tengyu Ma and Baishan Guo and Arpit Kalla and Markus Marks and Joseph Greer and Meng Wang and Peize Sun and Roman Rädle and Triantafyllos Afouras and Effrosyni Mavroudi and Katherine Xu and Tsung-Han Wu and Yu Zhou and Liliane Momeni and Rishi Hazra and Shuangrui Ding and Sagar Vaze and Francois Porcher and Feng Li and Siyuan Li and Aishwarya Kamath and Ho Kei Cheng and Piotr Dollár and Nikhila Ravi and Kate Saenko and Pengchuan Zhang and Christoph Feichtenhofer},
      year={2026},
      eprint={2511.16719},
      archivePrefix={arXiv},
      primaryClass={cs.CV},
      url={https://arxiv.org/abs/2511.16719}, 
}

@article{li2026mvsam3d,
  title={MV-SAM3D: Adaptive Multi-View Fusion for Layout-Aware 3D Generation},
  author={Li, Baicheng and Wu, Dong and Li, Jun and Zhou, Shunkai and Zeng, Zecui and Li, Lusong and Zha, Hongbin},
  journal={arXiv preprint arXiv:2603.11633},
  year={2026}
}

@InProceedings{foundationposewen2024,
    author    = {Wen, Bowen and Yang, Wei and Kautz, Jan and Birchfield, Stan},
    title     = {FoundationPose: Unified 6D Pose Estimation and Tracking of Novel Objects},
    booktitle = {Proceedings of the IEEE/CVF Conference on Computer Vision and Pattern Recognition (CVPR)},
    month     = {June},
    year      = {2024},
    pages     = {17868-17879}
}

@article{qiu2025lhmpp,
  title={LHM++: An Efficient Large Human Reconstruction Model for Pose-free Images to 3D},
  author={Lingteng Qiu and Peihao Li and Heyuan Li and Qi Zuo and Xiaodong Gu and Yuan Dong and Weihao Yuan and Rui Peng and Siyu Zhu and Xiaoguang Han and Guanying Chen and Zilong Dong},
  journal={arXiv preprint arXiv:2503.10625},
  year={2025}
}

@InProceedings{liu2025omniaudiogeneratingspatialaudio,
      title={OmniAudio: Generating Spatial Audio from 360-Degree Video}, 
      author={Huadai Liu and Tianyi Luo and Qikai Jiang and Kaicheng Luo and Peiwen Sun and Jialei Wan and Rongjie Huang and Qian Chen and Wen Wang and Xiangtai Li and Shiliang Zhang and Zhijie Yan and Zhou Zhao and Wei Xue},
      booktitle = {ICML},
      year={2025},
}

@inproceedings{mmaudio,
  title={{MMAudio}: Taming Multimodal Joint Training for High-Quality Video-to-Audio Synthesis},
  author={Cheng, Ho Kei and Ishii, Masato and Hayakawa, Akio and Shibuya, Takashi and Schwing, Alexander and Mitsufuji, Yuki},
  booktitle={CVPR},
  year={2025}
}

@inproceedings{visage, 
  title={ViSAGe: Video-to-Spatial Audio Generation}, 
  author={Jaeyeon Kim and Heeseung Yun and Gunhee Kim}, 
  booktitle={ICLR}, 
  year={2025}
}

@InProceedings{syncsde,
    author    = {Lee, Hyunjun and Lee, Hyunsoo and Han, Sookwan},
    title     = {SyncSDE: A Probabilistic Framework for Diffusion Synchronization},
    booktitle = {CVPR},
    year      = {2025}
}

@article{audiox,
  title={AudioX: Diffusion Transformer for Anything-to-Audio Generation},
  author={Tian, Zeyue and Jin, Yizhu and Liu, Zhaoyang and Yuan, Ruibin and Tan, Xu and Chen, Qifeng and Xue, Wei and Guo, Yike},
  journal={arXiv preprint arXiv:2503.10522},
  year={2025}
}

@misc{shan2025hunyuanvideofoleymultimodaldiffusionrepresentation,
      title={HunyuanVideo-Foley: Multimodal Diffusion with Representation Alignment for High-Fidelity Foley Audio Generation}, 
      author={Sizhe Shan and Qiulin Li and Yutao Cui and Miles Yang and Yuehai Wang and Qun Yang and Jin Zhou and Zhao Zhong},
      year={2025}
}

@article{multidiffusion,
  title={MultiDiffusion: Fusing Diffusion Paths for Controlled Image Generation},
  author={Bar-Tal, Omer and Yariv, Lior and Lipman, Yaron and Dekel, Tali},
  journal={arXiv preprint arXiv:2302.08113},
  year={2023}
}

@inproceedings{multifoley,
      author    = { Chen, Ziyang and 
                    Seetharaman, Prem and 
                    Russell, Bryan and 
                    Nieto, Oriol and
                    Bourgin, David and
                    Owens, Andrew and 
                    Salamon, Justin},
      title     = {Video-Guided Foley Sound Generation with Multimodal Controls},
      journal   = {CVPR},
      year      = {2025},
    }

@inproceedings{avdar,
    title = {Differentiable Room Acoustic Rendering with Multi-View Vision Priors},
    author = {Jin, Derong and Gao, Ruohan},
    booktitle = {ICCV},
    year = {2025},
}

@article{avr,
title={Acoustic Volume Rendering for Neural Impulse Response Fields},
author={Lan, Zitong and Zheng, Chenhao and Zheng, Zhiwei and Zhao, Mingmin},
journal={arXiv preprint arXiv:2411.06307},
year={2024}
}

@InProceedings{diffrir,
        title={Hearing Anything Anywhere},
        author={Mason Wang and Ryosuke Sawata and Samuel Clarke and Ruohan Gao and Shangzhe Wu and Jiajun Wu},
        booktitle={CVPR},
        year={2024}}

@inproceedings{liu2025haae,
  title = {Hearing Anywhere in Any Environment},
  author = {Liu, Xiulong and Kumar, Anurag and Calamia, Paul and Garí, Sebastià V. Amengual and Murdock, Calvin and Ananthabhotla, Ishwarya and Robinson, Philip and Shlizerman, Eli and Ithapu, Vamsi Krishna and Gao, Ruohan},
  booktitle = {CVPR},
  year = {2025},
}

@article{naf,
  title={Learning neural acoustic fields},
  author={Luo, Andrew and Du, Yilun and Tarr, Michael and Tenenbaum, Josh and Torralba, Antonio and Gan, Chuang},
  journal={NeurIPS},
  year={2022}
}

@inproceedings{soundspace2,
  title     =     {SoundSpaces 2.0: A Simulation Platform for Visual-Acoustic Learning},
  author    =     {Changan Chen and Carl Schissler and Sanchit Garg and Philip Kobernik and Alexander Clegg and Paul Calamia and Dhruv Batra and Philip W Robinson and Kristen Grauman},
  booktitle =     {NeurIPS},
  year      =     {2022}
}

@inproceedings{soundspace,
  title     =     {SoundSpaces: Audio-Visual Navigaton in 3D Environments},
  author    =     {Changan Chen and Unnat Jain and Carl Schissler and Sebastia Vicenc Amengual Gari and Ziad Al-Halah and Vamsi Krishna Ithapu and Philip Robinson and Kristen Grauman},
  booktitle =     {ECCV},
  year      =     {2020}
}

@inproceedings{genmo2025,
  title     = {GENMO: A GENeralist Model for Human MOtion},
  author    = {Li, Jiefeng and Cao, Jinkun and Zhang, Haotian and Rempe, Davis and Kautz, Jan and Iqbal, Umar and Yuan, Ye},
  booktitle = {ICCV},
  year      = {2025}
}

@article{bai2025recammaster,
  title={ReCamMaster: Camera-Controlled Generative Rendering from A Single Video},
  author={Bai, Jianhong and Xia, Menghan and Fu, Xiao and Wang, Xintao and Mu, Lianrui and Cao, Jinwen and Liu, Zuozhu and Hu, Haoji and Bai, Xiang and Wan, Pengfei and others},
  journal={arXiv preprint arXiv:2503.11647},
  year={2025}
}

@article{yang2026neoverse,
  author    = {Yang, Yuxue and Fan, Lue and Shi, Ziqi and Peng, Junran and Wang, Feng and Zhang, Zhaoxiang},
  title     = {NeoVerse: Enhancing 4D World Model with in-the-wild Monocular Videos},
  journal   = {arXiv preprint arXiv:2601.00393},
  year      = {2026},
}

@article{xie2026lavr,
  title={LaVR: Scene Latent Conditioned Generative Video Trajectory Re-Rendering using Large 4D Reconstruction Models},
  author={Xie, Mingyang and Khan, Numair and Wang, Tianfu and Dhingra, Naina and Nam, Seonghyeon and Yang, Haitao and Hui, Zhuo and Metzler, Christopher and Vedaldi, Andrea and Pirsiavash, Hamed and others},
  journal={arXiv preprint arXiv:2601.14674},
  year={2026}
}

@inproceedings{som2024,
  title     = {Shape of Motion: 4D Reconstruction from a Single Video},
  author    = {Wang, Qianqian and Ye, Vickie and Gao, Hang and Zeng, Weijia and Austin, Jake and Li, Zhengqi and Kanazawa, Angjoo},
  booktitle   = {ICCV},
  year      = {2025}
}

@inproceedings{priorenhancedgaussian,
author = {Shih, Meng-Li and Chen, Ying-Huan and Liu, Yu-Lun and Curless, Brian},
title = {Prior-Enhanced Gaussian Splatting for Dynamic Scene Reconstruction from Casual Video},
year = {2025},
booktitle = {SIGGRAPH Asia}
}

@article{pad3r,
    author    = {Liao, Ting-Hsuan and Liu, Haowen and Xu, Yiran and Ge, Songwei and Yang, Gengshan and Huang, Jia-Bin},
    title     = {PAD3R: Pose-Aware Dynamic 3D Reconstruction from Casual Videos},
    journal   = {SIGGRAPH ASIA},
    year      = {2025},
  }

@InProceedings{ActionMesh2026,
    author    = {Sabathier, Remy and Novotny, David and Mitra, Niloy J. and Monnier, Tom},
    title     = {ActionMesh: Animated 3D Mesh Generation with Temporal 3D Diffusion},
    booktitle = {Proceedings of the IEEE/CVF Conference on Computer Vision and Pattern Recognition (CVPR)},
    month     = {June},
    year      = {2026},
    pages     = {34312-34321}
}

@InProceedings{chen2026motion3to4,
    author    = {Chen, Hongyuan and Chen, Xingyu and Xu, Zexiang and Chen, Anpei},
    title     = {Motion 3-to-4: 3D Motion Reconstruction for 4D Synthesis},
    booktitle = {Proceedings of the IEEE/CVF Conference on Computer Vision and Pattern Recognition (CVPR)},
    month     = {June},
    year      = {2026},
    pages     = {28947-28958}
}

@InProceedings{chen2025v2m4,
    author    = {Chen, Jianqi and Zhang, Biao and Tang, Xiangjun and Wonka, Peter},
    title     = {V2M4: 4D Mesh Animation Reconstruction from a Single Monocular Video},
    booktitle = {ICCV},
    year      = {2025}
}

@inproceedings{hu2024gaussianavatar,
    title={GaussianAvatar: Towards Realistic Human Avatar Modeling from a Single Video via Animatable 3D Gaussians},
    author={Hu, Liangxiao and Zhang, Hongwen and Zhang, Yuxiang and Zhou, Boyao and Liu, Boning and Zhang, Shengping and Nie, Liqiang},
    booktitle={CVPR},
    year={2024}
}

@inproceedings{moon2024exavatar,
  title={Expressive Whole-Body 3D Gaussian Avatar},
  author = {Moon, Gyeongsik and Shiratori, Takaaki and Saito, Shunsuke},  
  booktitle={ECCV},
  year={2024}
}

@inproceedings{
    syncdiffusion,
    title={SyncDiffusion: Coherent Montage via Synchronized Joint Diffusions},
    author={Yuseung Lee and Kunho Kim and Hyunjin Kim and Minhyuk Sung},
    booktitle={NeurIPS},
    year={2023},
}

@inproceedings{gao2019visual-sound,
  title = {2.5D Visual Sound},
  author = {Gao, Ruohan and Grauman, Kristen},
  booktitle = {CVPR},
  year = {2019},
}

@inproceedings{garg2023visually,
  title = {Visually-Guided Audio Spatialization in Video with Geometry-Aware Multi-task Learning},
  author = {Garg, Rishabh and Gao, Ruohan and Grauman, Kristen},
  booktitle = {IJCV},
  year = {2023},
}

@inproceedings{garg2021geometry,
  title={Geometry-Aware Multi-Task Learning for Binaural Audio Generation from Video},
  author={Garg, Rishabh and Gao, Ruohan and Grauman, Kristen},
  booktitle={BMVC},
  year={2021}
}

@article{morgado2018self,
  title={Self-supervised generation of spatial audio for 360 video},
  author={Morgado, Pedro and Nvasconcelos, Nuno and Langlois, Timothy and Wang, Oliver},
  journal={NeurIPS},
  year={2018}
}

@article{li2018scene,
  title={Scene-aware audio for 360 videos},
  author={Li, Dingzeyu and Langlois, Timothy R and Zheng, Changxi},
  journal={TOG},
  year={2018}
}

@inproceedings{av-nerf,
 author = {Liang, Susan and Huang, Chao and Tian, Yapeng and Kumar, Anurag and Xu, Chenliang},
 booktitle = {NeurIPS},
 title = {AV-NeRF: Learning Neural Fields for Real-World Audio-Visual Scene Synthesis},
 year = {2023}
}

@article{chen2024images,
  title     = {Images that Sound: Composing Images and Sounds on a Single Canvas},
  author    = {Chen, Ziyang and Geng, Daniel and Owens, Andrew},
  journal = {NeurIPS},
  year      = {2024}
}

@inproceedings{SMPL-X:2019,
    title = {Expressive Body Capture: 3D Hands, Face, and Body from a Single Image},
    author = {Pavlakos, Georgios and Choutas, Vasileios and Ghorbani, Nima and Bolkart, Timo and Osman, Ahmed A. A. and Tzionas, Dimitrios and Black, Michael J.},
    booktitle = {CVPR},
    year = {2019}
}

@inproceedings{geocalib,
  author    = {Alexander Veicht and
               Paul-Edouard Sarlin and
               Philipp Lindenberger and
               Marc Pollefeys},
  title     = {{GeoCalib: Single-image Calibration with Geometric Optimization}},
  booktitle = {ECCV},
  year      = {2024}
}

@misc{worldlabs_marble,
  author = {World Labs},
  title = {Marble},
  year = {2025},
  howpublished = {\url{https://marble.worldlabs.ai/}},
}

@misc{nomadoor2026flux2panorama,
  author       = {nomadoor},
  title        = {{FLUX.2 Klein 9B 360 ERP Outpaint LoRA}},
  year         = {2026},
  howpublished = {\url{https://huggingface.co/nomadoor/flux-2-klein-9B-360-erp-outpaint-lora}},
  note         = {Hugging Face model repository}
}

@misc{flux2,
    author={Black Forest Labs},
    title={{FLUX.2: Frontier Visual Intelligence}},
    year={2025},
    howpublished={\url{https://bfl.ai/blog/flux-2}},
}

@article{sam3dteam2025sam3d3dfyimages,
      title={SAM 3D: 3Dfy Anything in Images}, 
      author={SAM 3D Team and Xingyu Chen and Fu-Jen Chu and Pierre Gleize and Kevin J Liang and Alexander Sax and Hao Tang and Weiyao Wang and Michelle Guo and Thibaut Hardin and Xiang Li and Aohan Lin and Jiawei Liu and Ziqi Ma and Anushka Sagar and Bowen Song and Xiaodong Wang and Jianing Yang and Bowen Zhang and Piotr Dollár and Georgia Gkioxari and Matt Feiszli and Jitendra Malik},
      year={2025},
      eprint={2511.16624},
      archivePrefix={arXiv},
      primaryClass={cs.CV},
      url={https://arxiv.org/abs/2511.16624}, 
}

@inproceedings{wang2026vggtomega,
  title     = {{VGGT-$\Omega$}},
  author    = {Jianyuan Wang and Minghao Chen and Shangzhan Zhang and Nikita Karaev and Johannes Sch{\"o}nberger and Patrick Labatut and Piotr Bojanowski and David Novotny and Andrea Vedaldi and Christian Rupprecht},
  booktitle = {Proceedings of the IEEE/CVF Conference on Computer Vision and Pattern Recognition (CVPR)},
  year      = {2026}
}

@article{team2026kimi,
  title={Kimi k3: Open frontier intelligence},
  author={Team, Kimi and Bai, Tongtong and Bai, Yifan and Bao, Yiping and Cai, Jianfeng and Cai, Xinyuan and Cao, Peizhou and Cao, Yuxuan and Chai, Ziwei and Charles, Y and others},
  journal={arXiv preprint arXiv:2607.24653},
  year={2026}
}

@article{hyworld2026,
  title={HY-World 2.0: A Multi-Modal World Model for Reconstructing, Generating, and Simulating 3D Worlds},
  author={{Team HY-World}},
  journal={arXiv preprint arXiv:2604.14268},
  year={2026},
  doi={10.48550/arXiv.2604.14268},
  url={https://arxiv.org/abs/2604.14268}
}

@article{yoshida2026scene2sound,
  title={Scene2Sound: Auditory-Grounded Soundscape Generation for 3D Gaussian Worlds},
  author={Yoshida, Masaki and Togo, Ren and Ogawa, Takahiro and Haseyama, Miki},
  journal={arXiv preprint arXiv:2608.00463},
  year={2026}
}

@article{litman2026lift4d,
  author  = {Litman, Yehonathan and Ma, Xiaoxuan and Shah, Manan and Ugrinovic, Nicol\'{a}s and Kitani, Kris and De la Torre, Fernando and Tulsiani, Shubham},
  title   = {Lift4D: Harmonizing Single-View 3D Estimation for 4D Reconstruction In-the-Wild},
  journal = {arXiv preprint arXiv:2606.23688},
  year    = {2026},
}

@article{helix4d2026,
  title={Helix4D: Complex 4D Mesh Generation},
  author={Yenphraphai, Jiraphon and Chen, Jianqi and Wang, Jian and Qian, Gordon Guocheng and Tulyakov, Sergey and Abdal, Rameen and Yeh, Raymond A. and Wonka, Peter and Wang, Chaoyang},
  journal={arXiv preprint arXiv:2605.26109},
  year={2026}
}

@inproceedings{magnenat1988joint,
author = {Magnenat-Thalmann, N. and Laperri{\`e}re, R. and Thalmann, D.},
title = {Joint-dependent local deformations for hand animation and object grasping},
year = {1989},
publisher = {Canadian Information Processing Society},
address = {CAN},
booktitle = {Proceedings on Graphics Interface '88},
pages = {26–33},
numpages = {8},
location = {Edmonton, Alberta, Canada}
}

@inproceedings{chen2026audible,
  title     = {Audible World Models: Spatially Aware Sound Generation for 3D Worlds},
  author    = {Chen, Duowen and He, Jinjin and KV, Gouthaman and Bangalore Venkatesh, Sandeep and Zhu, Bo},
  booktitle = {Advances in Neural Information Processing Systems (NeurIPS)},
  year      = {2026}
}

@misc{omnidream,
      title={Enabling Immersive Audio-Visual Experience from Any Video}, 
      author={Zitong Lan and Mutian Tong and Jiatao Gu and Mingmin Zhao},
      year={2026},
      eprint={2609.36295},
      archivePrefix={arXiv},
      primaryClass={cs.SD},
      url={https://arxiv.org/abs/2609.36295}, 
}

@article{traer2016statistics,
  title   = {Statistics of natural reverberation enable perceptual
             separation of sound and space},
  author  = {Traer, James and McDermott, Josh H.},
  journal = {Proceedings of the National Academy of Sciences},
  volume  = {113},
  number  = {48},
  pages   = {E7856--E7865},
  year    = {2016},
  doi     = {10.1073/pnas.1612524113}
}

@article{traer2021causal,
  title   = {Causal inference in environmental sound recognition},
  author  = {Traer, James and Norman-Haignere, Sam V. and
             McDermott, Josh H.},
  journal = {Cognition},
  volume  = {214},
  pages   = {104627},
  year    = {2021},
  doi     = {10.1016/j.cognition.2021.104627}
}
\bibliographystyle{main}

\clearpage
\newpage
\appendix

\section*{Appendix}

\section{Motion-Type-Specific Instantiations for Dynamic Foreground Generation}
\label{app:motion_instantiations}
\begin{figure*}[t]
    \centering
    \includegraphics[width=\textwidth]{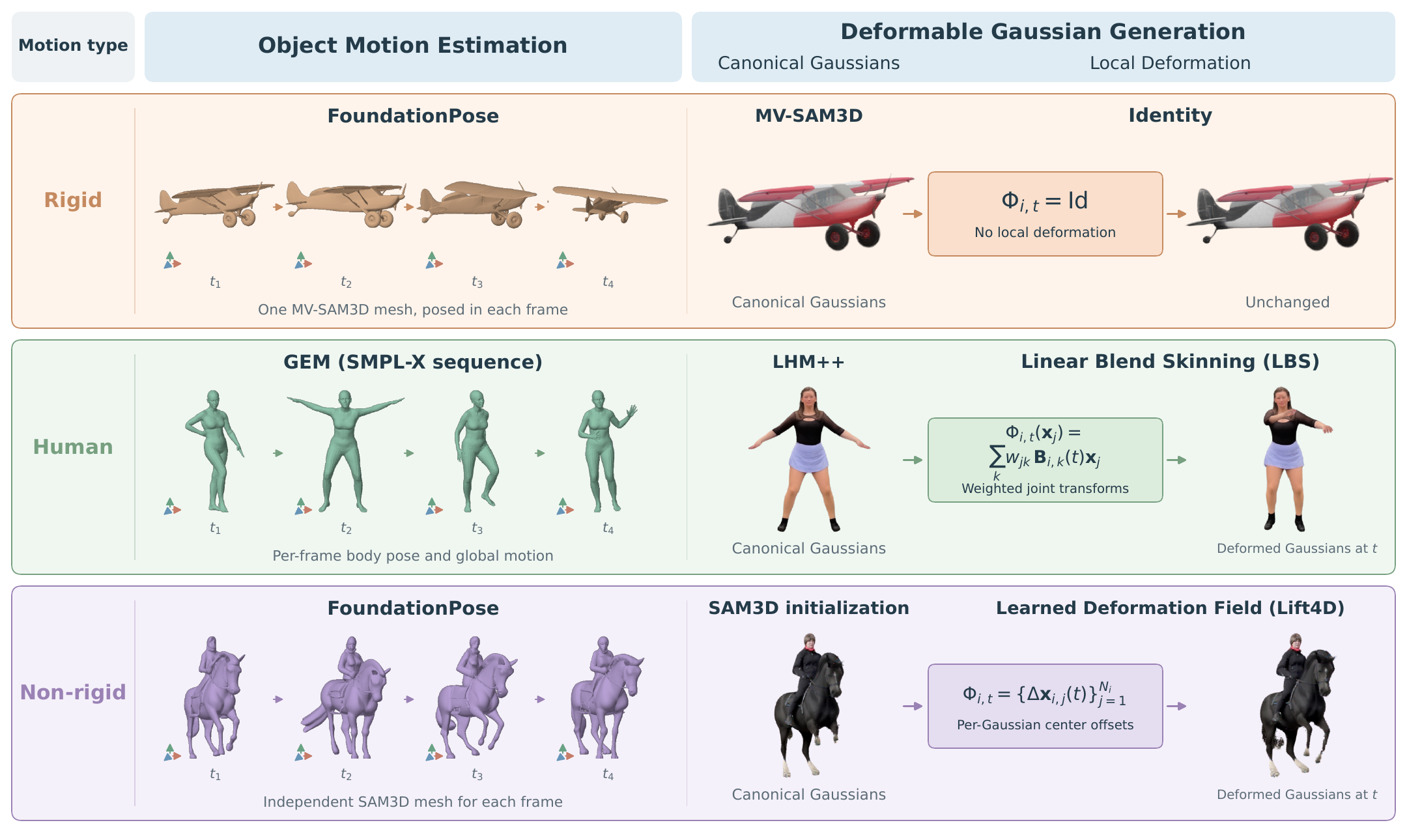}
    \vspace{-1.5em}
    \caption{
    \textbf{Motion-type-specific instantiations for dynamic foreground generation.} Each branch combines object motion estimation with canonical Gaussian generation and a local deformation model: identity for rigid objects, linear blend skinning~\citep{magnenat1988joint} for humans, and learned per-Gaussian center offsets~\citep{litman2026lift4d} for non-rigid objects.
    }
    \label{fig:motion_instantiations}
\end{figure*}

Our dynamic foreground generation follows the same two
components for all motion types:
\textit{Object Motion Estimation}, which estimates the
global similarity transformation
$\mathbf{H}_{i,t}^{\mathrm{obj}\to\mathrm{cam}}$
from object space to camera space at each frame, and
\textit{Deformable Gaussian Generation}, which produces
canonical Gaussians $\mathcal{G}_i^{\mathrm{obj}}$ and their
time-dependent deformation $\Phi_{i,t}$.
We instantiate these components with different backbones
for rigid-body, human, and general deformable motion,
while retaining the shared formulation:
\begin{equation}
\mathcal{G}_{i,t}^{\mathrm{world}}
=
\mathbf{T}_{t}^{\mathrm{c}\to\mathrm{w}}
\left[
\left(
\mathbf{H}_{i,t}^{\mathrm{obj}\to\mathrm{cam}}
\circ \Phi_{i,t}
\right)
\left(\mathcal{G}_i^{\mathrm{obj}}\right)
\right].
\end{equation}
Fig.~\ref{fig:motion_instantiations} summarizes how
each motion type instantiates these two components. All outputs are subsequently aligned
with the panoramic background through the
foreground-background alignment described in the main paper.

\subsection{Rigid-Body Motion}
\label{app:rigid_motion}

\noindent\textbf{Object Motion Estimation.}
We use MV-SAM3D~\citep{li2026mvsam3d} to jointly
reconstruct a proxy mesh and canonical Gaussians
from selected frames of the input video.
Using this mesh,
FoundationPose~\citep{foundationposewen2024}
estimates object rotation and translation through
anchor-frame registration and temporal tracking.
We complete motion estimates across unreliable
or missing observations using trajectory interpolation
and extrapolation.
Together with reconstruction-derived scale estimates,
the resulting poses specify
$\mathbf{H}_{i,t}^{\mathrm{obj}\to\mathrm{cam}}$.
Keyframe selection, tracking, and trajectory completion
are detailed in
Appendix~\ref{app:foreground_implementation}.

\noindent\textbf{Deformable Gaussian Generation.}
We use the canonical Gaussians $\mathcal{G}_i^{\mathrm{obj}}$
jointly produced with the proxy mesh by MV-SAM3D.
Reconstruction is conditioned on foreground images,
masks, and depth-derived pointmaps from selected
keyframes.
Since rigid objects preserve their shape,
we set $\Phi_{i,t}=\mathrm{Id}$ and reuse the same
canonical Gaussians throughout the sequence.

\subsection{Human Motion}
\label{app:human_motion}

\noindent\textbf{Object Motion Estimation.}
We use GEM ~\citep{genmo2025} to recover SMPL-X ~\citep{SMPL-X:2019} parameters from the
foreground video, including global body motion and
articulated poses. The global body pose provides
the rotation and translation needed for
$\mathbf{H}_{i,t}^{\mathrm{obj}\to\mathrm{cam}}$.
We reconcile the human reconstruction with the scene
scale using the ratio between predicted scene depth
and rendered human depth over foreground pixels.
The camera poses from video preprocessing then map
the recovered motion into the common world frame.

\noindent\textbf{Deformable Gaussian Generation.}
We use LHM++ ~\citep{qiu2025lhmpp} to reconstruct an animatable Gaussian
avatar from reference images of the person.
The avatar provides $\mathcal{G}_i^{\mathrm{obj}}$, and skeletal
skinning driven by the estimated SMPL-X pose sequence
instantiates $\Phi_{i,t}$. This captures body articulation
while maintaining a shared appearance representation
across frames. 

\subsection{General Deformable Motion}
\label{app:general_deformable_motion}

\noindent\textbf{Object Motion Estimation.}
We use SAM3D~\citep{sam3dteam2025sam3d3dfyimages}
to reconstruct a proxy mesh for each frame,
conditioned on the foreground image, mask,
and depth-derived pointmap.
We use these per-frame meshes with
FoundationPose~\citep{foundationposewen2024}
to estimate object rotation and translation
through pose registration and tracking.
Allowing the proxy geometry to vary across frames
accounts for the object's changing shape.
Together with reconstruction-derived scale estimates,
the resulting poses specify
$\mathbf{H}_{i,t}^{\mathrm{obj}\to\mathrm{cam}}$.

\noindent\textbf{Deformable Gaussian Generation.}
We initialize the canonical Gaussians
$\mathcal{G}_i^{\mathrm{obj}}$ from the SAM3D reconstruction
of a reference frame and use
Lift4D~\citep{litman2026lift4d} to fit the time-dependent
deformation field $\Phi_{i,t}$ to the per-frame
reconstructions and observed foreground images.
The deformation field is parameterized by learned
control nodes, supporting non-human articulation
and flexible shape changes without assuming a human
body model. We combine the deformed Gaussians with
the estimated object-to-camera transformations using
the same formulation as the other two instantiations.

\section{Details on Foreground-Background Alignment}
\label{app:fg_bg_alignment}

Foreground reconstruction and panorama-to-3D generation
produce geometry independently. Although their coordinate
orientations are established by camera estimation and
panorama calibration, their scales and local surfaces
may disagree. We address these discrepancies through
a scene-level background scale followed by time-varying
foreground displacement corrections.

\subsection{Global Scale Alignment}

We render the video-derived background point cloud and
the generated background Gaussians into the same
panoramic view.
Let $d_{\mathrm{init}}(u)$ and $d_{\mathrm{bg}}(u)$
denote their radial depths at panorama pixel $u$,
respectively.
We retain reliable correspondences over their common
visible region, denoted by $\mathcal{U}$, and weight
them by observation confidence
$w_u=C_{\mathrm{obs}}(u)$.

We estimate a single scene-level background scale
through weighted least squares:
\begin{equation}
s
=
\underset{\tilde{s}>0}{\arg\min}
\sum_{u\in\mathcal{U}}
w_u
\left(
\tilde{s}\,d_{\mathrm{bg}}(u)-d_{\mathrm{init}}(u)
\right)^2
=
\frac{
\sum_{u\in\mathcal{U}}
w_u\,d_{\mathrm{bg}}(u)d_{\mathrm{init}}(u)
}{
\sum_{u\in\mathcal{U}}
w_u\,d_{\mathrm{bg}}(u)^2
}.
\label{eq:bg_scale_alignment}
\end{equation}
We apply this scale to the generated background geometry
and map it into the common world frame using the known
panorama-to-world transformation.

\subsection{Local Displacement Estimation}

Global scale alignment does not resolve local geometric
differences between the two reconstructions.
For each object $i$ and frame $t$, we collect background
points in a neighborhood of its foreground mask
$M_{i,t}$. We exclude foreground pixels and use depth
and surface-normal discontinuities to restrict the
neighborhood to connected geometry.
For supported objects, such as a car on a road,
this neighborhood captures the nearby supporting surface.

We project these points into the reference panorama
and retain correspondences that are visible and
depth-consistent with the initialized background,
with valid geometry in the generated background.
For each retained correspondence $p$, let
$\mathbf{x}_{i,t,p}$ be the video-derived point and
$\mathbf{y}_{i,t,p}$ its corresponding point on the
scale-aligned background, both in world space.
Their displacement is
$\Delta\mathbf{x}_{i,t,p}
=\mathbf{y}_{i,t,p}-\mathbf{x}_{i,t,p}$.

After rejecting displacement outliers, we estimate
the local translation by a weighted average:
\begin{equation}
\widehat{\delta}_{i,t}
=
\frac{
\sum_{p\in\Omega_{i,t}}
w_{i,t,p}\,\Delta\mathbf{x}_{i,t,p}
}{
\sum_{p\in\Omega_{i,t}}w_{i,t,p}
},
\label{eq:local_alignment_translation}
\end{equation}
where $\Omega_{i,t}$ is the retained support set and
$w_{i,t,p}$ combines depth confidence with background
rendering opacity. We accept a displacement estimate
only when its support is sufficient and its residual
displacements are consistent.

\subsection{Displacement Interpolation}

Frames without reliable local support, such as an
airborne interval, do not provide direct displacement
constraints. We propagate the accepted estimates
through a temporally regularized correction field.
Let $\mathcal{T}_i$ denote the supported frames and
$c_{i,t}$ the confidence of each accepted estimate.
We solve
\begin{equation}
\min_{\{\delta_{i,t}\}_{t=1}^{T}}
\sum_{t\in\mathcal{T}_i}
c_{i,t}\,
\rho\!\left(
\|\delta_{i,t}-\widehat{\delta}_{i,t}\|_2
\right)
+
\eta\sum_{t=1}^{T-1}
\|\delta_{i,t+1}-\delta_{i,t}\|_2^2,
\label{eq:alignment_temporal_propagation}
\end{equation}
where $\rho$ is a Huber penalty and $\eta$ controls
temporal smoothness. 

This optimization smooths the supported displacement
estimates and linearly interpolates the resulting
offsets across intervals without reliable contact.
Outside the supported temporal range, the correction
remains equal to the nearest supported offset.
If no reliable support is available throughout the
sequence, we set $\delta_{i,t}=\mathbf{0}$ and retain
the original trajectory.

Finally, we update each object's world-space position as
\begin{equation}
\widetilde{\mathbf{p}}_{i,t}^{\mathrm{world}}
=
\mathbf{p}_{i,t}^{\mathrm{world}}+\delta_{i,t},
\end{equation}
and apply the same translation to its Gaussians,
preserving object rotations and local deformations.
Together with the scale-aligned background, these
corrected foreground objects form the aligned
4D visual scene.

\section{Ablation on Visual Components}
\label{app:visual_ablation}

\begin{figure}[t]
    \centering
    \begin{minipage}[t]{0.49\linewidth}
        \centering
        \includegraphics[width=\linewidth]{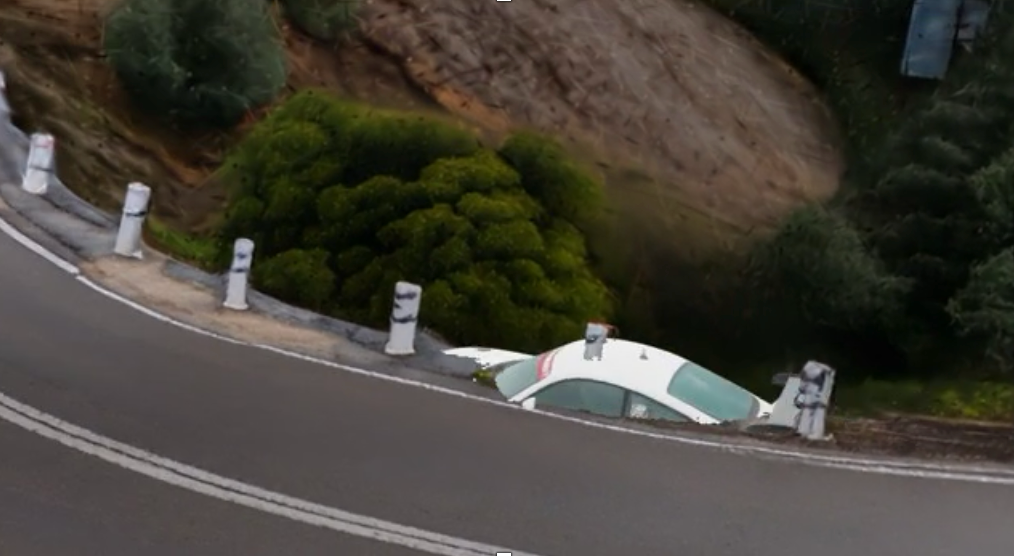}
        \par\smallskip
        {\small (a) Scale only}
    \end{minipage}\hfill
    \begin{minipage}[t]{0.49\linewidth}
        \centering
        \includegraphics[width=\linewidth]{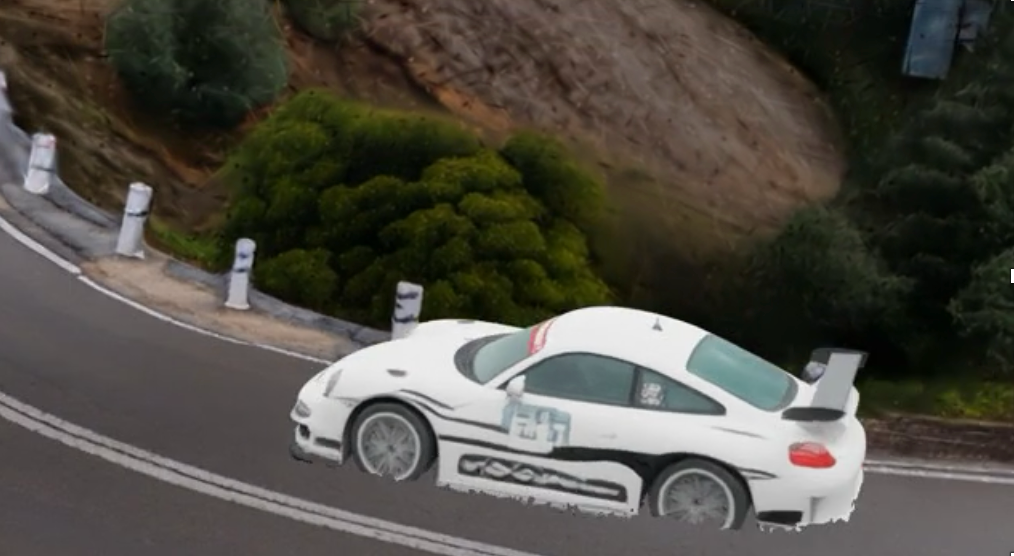}
        \par\smallskip
        {\small (b) Scale + displacement}
    \end{minipage}

    \medskip

    \begin{minipage}[t]{0.49\linewidth}
        \centering
        \includegraphics[width=\linewidth]{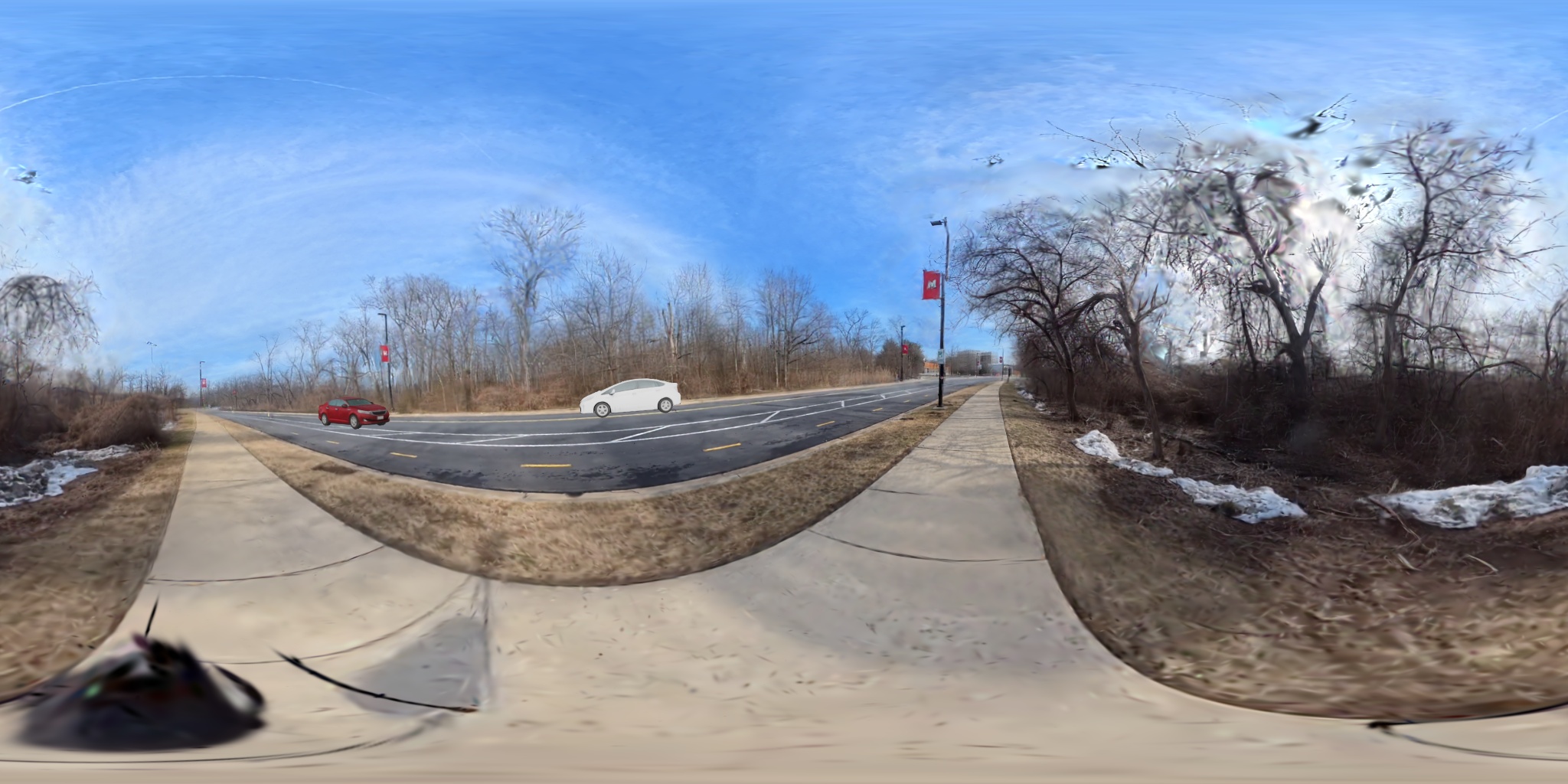}
        \par\smallskip
        {\small (c) HYWorld 2.0}
    \end{minipage}\hfill
    \begin{minipage}[t]{0.49\linewidth}
        \centering
        \includegraphics[width=\linewidth]{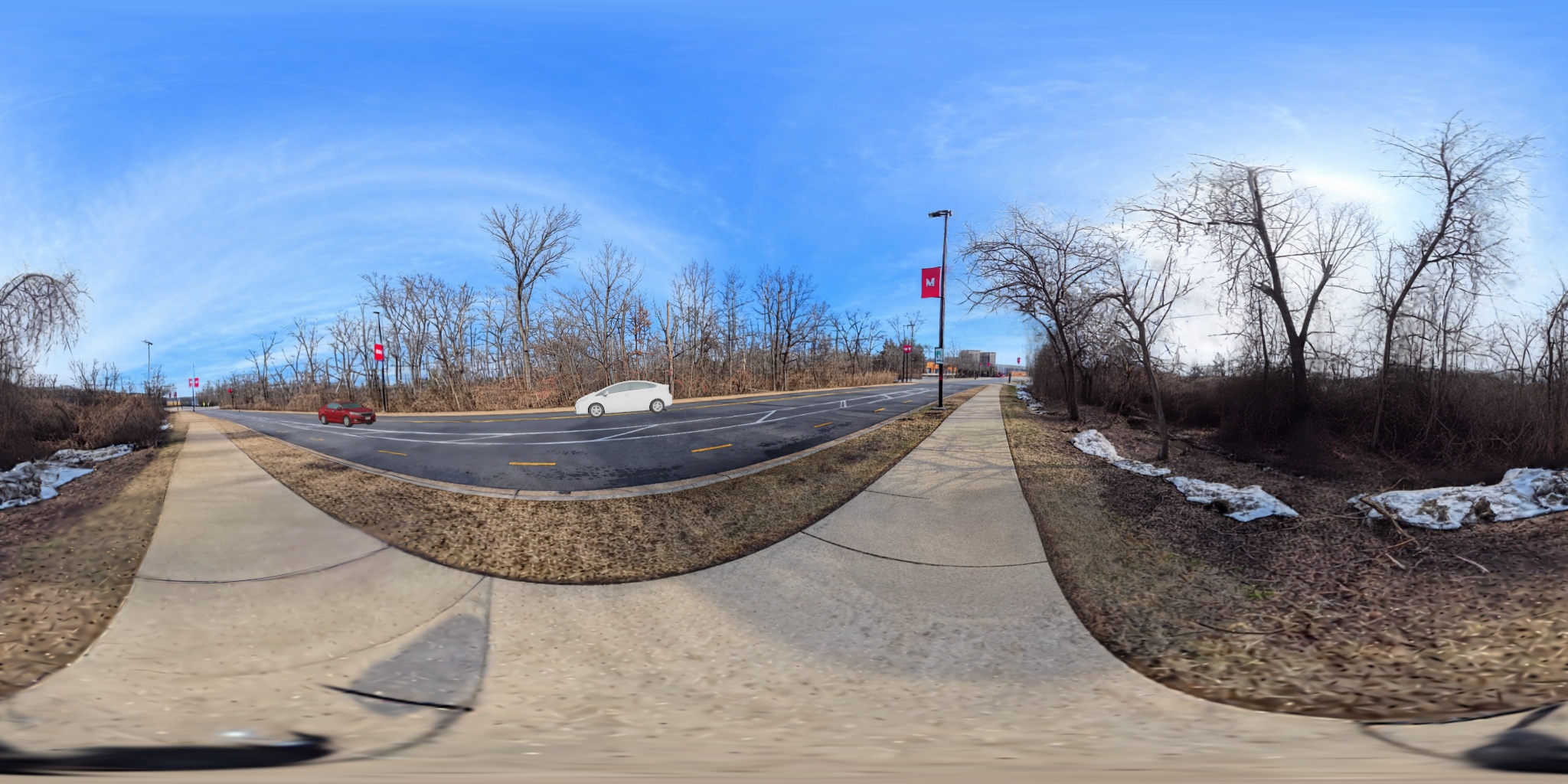}
        \par\smallskip
        {\small (d) Marble}
    \end{minipage}

    \caption{
    \textbf{Qualitative ablation on visual components.} \textbf{Top:} Ablation of displacement correction in
foreground-background alignment.
Both variants include global scale alignment;
adding local displacement corrections reduces
foreground interpenetration with the ground.
    \textbf{Bottom:} Both panorama-to-3D backbones produce
    usable scenes, but Marble yields fewer visible
    reconstruction artifacts in the illustrated example.
    }
    \label{fig:visual_ablation}
\end{figure}

We examine two components of visual scene generation:
the displacement correction in foreground-background
alignment and the panorama-to-3D backbone used for
background generation. Fig.~\ref{fig:visual_ablation}
provides qualitative comparisons.

\noindent\textbf{Foreground-background displacement correction.}
We compare our full alignment, which combines global
scale alignment with local displacement corrections,
against a variant that retains only scale alignment.
Global scale alone cannot resolve local geometric
discrepancies between the independently reconstructed
foreground and background. The scale-only variant
places much of the car below the road surface.
Adding displacement corrections substantially reduces
this interpenetration and improves object-ground contact.
This comparison illustrates the importance of accounting
for local geometry in addition to global scale.

\noindent\textbf{Panorama-to-3D backbone.}
We compare HYWorld 2.0~\citep{hyworld2026} and Marble~\citep{worldlabs_marble} for lifting the
completed background panorama into 3D Gaussians.
Both produce usable background reconstructions,
demonstrating that our framework accommodates different
panorama-to-3D backbones. In our qualitative comparisons,
however, HYWorld 2.0 exhibits more visible artifacts,
including distorted vegetation and spurious geometry,
while Marble produces cleaner renderings.
For reproducibility, we use HYWorld 2.0 for all
quantitative evaluations. For improved visual quality,
we use the proprietary Marble model in our qualitative
comparisons and demos.

\section{Forward and Inverse Acoustic Rendering}
\label{app:forward_inverse_acoustic}

This section details the forward renderer $\mathcal{R}_i$ and inverse renderer $\mathcal{Q}_i$ used for diffusion synchronization, and explains how occlusion is handled during camera-to-object fusion.

\paragraph{Direct-Path Propagation.}
Measurements of natural acoustics show that reverberation typically decays faster outdoors than indoors~\citep{traer2016statistics}. Perceptual experiments further show that strong room-like reverberation can sound atypical for low-intensity outdoor sources~\citep{traer2021causal}. Motivated by these observations, we adopt an outdoor-focused direct-path approximation and omit reflected paths and reverberation. Our renderer accounts for Doppler shifts, distance attenuation, and air absorption, which change the timing and amplitude of each source signal.

For source $i$, let $t$ denote reception time at the camera and $u$ denote emission time in object space. We write the emission-time mapping as $u=\phi_i(t)=t_i^{\mathrm{em}}(t)$. This mapping is determined by the source and camera trajectories, and its local rate accounts for the Doppler shift. Let $g_i(t)>0$ denote the multiplicative gain accounting for distance attenuation and air absorption. On available, unoccluded reception intervals $\Omega_i$, the forward renderer is
\begin{equation}
    \bigl(\mathcal{R}_i a_i^{\mathrm{obj}}\bigr)(t)
    =
    g_i(t)\,
    a_i^{\mathrm{obj}}\!\left(\phi_i(t)\right),
    \qquad t\in\Omega_i.
    \label{eq:app_forward_rendering}
\end{equation}
This expression describes the unoccluded direct path. Occluded intervals are excluded from inverse-rendering feedback, as described below.

\paragraph{Inverse Rendering.}
Given the source and camera trajectories, we treat $\phi_i$ and $g_i$ as known. We assume that $\phi_i$ is strictly increasing over the time range considered, so that each emission time has a unique corresponding reception time. The inverse renderer first reverses this time mapping and then divides out the propagation gain. For a separated camera-space signal $b_i$,
\begin{equation}
    \bigl(\mathcal{Q}_i b_i\bigr)(u)
    =
    \frac{
        b_i\!\left(\phi_i^{-1}(u)\right)
    }{
        g_i\!\left(\phi_i^{-1}(u)\right)
    },
    \qquad u\in\mathcal{U}_i,
    \label{eq:app_inverse_rendering}
\end{equation}
where $\mathcal{U}_i=\phi_i(\Omega_i)$ is the corresponding set of valid emission times. Substituting Eq.~\ref{eq:app_forward_rendering} into Eq.~\ref{eq:app_inverse_rendering} gives
\begin{equation}
    \begin{aligned}
        \bigl(\mathcal{Q}_i\mathcal{R}_i
        a_i^{\mathrm{obj}}\bigr)(u)
        &=
        \frac{
            g_i\!\left(\phi_i^{-1}(u)\right)
            a_i^{\mathrm{obj}}\!\left(
                \phi_i\!\left(\phi_i^{-1}(u)\right)
            \right)
        }{
            g_i\!\left(\phi_i^{-1}(u)\right)
        }
        \\
        &= a_i^{\mathrm{obj}}(u),
        \qquad u\in\mathcal{U}_i.
    \end{aligned}
    \label{eq:app_inverse_identity}
\end{equation}
Thus, the continuous-time propagation model is analytically invertible on valid intervals when the time mapping is strictly increasing and the gain is nonzero. In discrete time, evaluating the warped signal requires
resampling, which introduces numerical approximation errors.
Source-separation errors further affect the object-space estimate obtained from a camera-space mixture.

\paragraph{Occlusion Handling and Fusion.}
We do not apply inverse rendering to occluded intervals. Instead, we define a validity mask on the object-space timeline:
\begin{equation}
    m_i(u)
    =
    \begin{cases}
        1, & u\in\mathcal{U}_i,\\
        0, & \text{otherwise}.
    \end{cases}
    \label{eq:app_validity_mask}
\end{equation}
The mask associates each emission time with the visibility and availability
of its corresponding camera-space observation.

At diffusion step $\tau$, we obtain the inverse target only on valid
intervals:
\begin{equation}
    \tilde{a}_{i,\tau}^{\mathrm{obj}}(u)
    =
    \left[
        \mathcal{Q}_i\!\left(
            \operatorname{Sep}_{q_i}\!\left(
                \hat{a}_{\tau}^{\mathrm{cam}}
            \right)
        \right)
    \right](u),
    \qquad u\in\mathcal{U}_i.
    \label{eq:app_inverse_target}
\end{equation}
The object-space fusion is then
\begin{equation}
    \bar{a}_{i,\tau}^{\mathrm{obj}}(u)
    =
    \begin{cases}
        (1-\eta_{\mathrm{obj}})
        \hat{a}_{i,\tau}^{\mathrm{obj}}(u)
        +
        \eta_{\mathrm{obj}}
        \tilde{a}_{i,\tau}^{\mathrm{obj}}(u),
        & m_i(u)=1,\\[3pt]
        \hat{a}_{i,\tau}^{\mathrm{obj}}(u),
        & m_i(u)=0.
    \end{cases}
    \label{eq:app_masked_fusion}
\end{equation}
Visible intervals therefore receive camera-to-object feedback, while occluded or unavailable intervals retain the object-space branch's own prediction during fusion. This avoids attempting to recover source signals from intervals where their camera-space contributions cannot be reliably inverted.

\section{Details on Our \ourdataset Dataset}
\label{app:dataset_details}

We collect \ourdataset to evaluate spatial audio
generation at listener positions distinct from the
input camera viewpoints. The dataset contains
54 dynamic scenes: 25 vehicle scenes and 29 human
scenes, covering rigid-body motion and human
articulation. Each scene includes a panoramic video
and synchronized first-order ambisonic (FOA) audio
recorded at a separate listening position.
The dataset includes both single-source and
multi-source scenarios, with 13 scenes containing
two dynamic foreground sound sources.

\noindent\textbf{Vehicle scenes.}
We capture 25 vehicle scenes across three locations,
featuring police vehicles, pickup trucks, sedans,
and buses. These scenes cover different vehicle
appearances and acoustic characteristics, with
vehicles moving through the scene along different
trajectories. Three scenes contain two vehicles,
requiring the generated audio to account for multiple
moving sources and their respective spatial cues.
The remaining 22 scenes contain a single foreground
vehicle.

\noindent\textbf{Human scenes.}
We capture 29 human scenes across three locations.
Activities include walking, jogging, jumping,
foot stomping, clapping, burpees, and clapping while
walking. These activities produce footsteps,
hand claps, and ground-contact sounds with distinct
temporal patterns, enabling evaluation of both spatial audio and synchronization with visible actions. Ten scenes contain two people performing either the same activity or different activities, such as walking and jogging, or clapping and jogging. The remaining 19 scenes contain one person. Some recordings additionally include static background sounds, such as flowing water, providing mixtures of dynamic foreground events and persistent background audio.

Together, these scenes cover diverse source motions, acoustic events, and overlapping sound sources. The FOA recordings provide reference audio at a listening
position different from the input camera position,
allowing us to evaluate spatial audio generation
at novel listener positions. We follow the annotation and capture-device setup
protocols of the \textsc{SonoScene360} dataset in SonoWorld~\citep{sonoworld}.

For experiments on \ourdataset, we use our captured
full background panorama as input to panorama-to-3D
generation. For each dynamic foreground sound source,
we extract a perspective field-of-view (FoV) video
from the recorded panoramic sequence using mask-based
tracking. Specifically, per-frame foreground masks
determine the viewing direction of a virtual camera,
whose orientation is temporally smoothed to follow
the source with a fixed FoV. We project the panoramic
frames into these views to obtain the foreground video
inputs to our method, producing one tracked FoV video
per dynamic sound source.

\section{Details on User Study}
\label{app:user_study}

\noindent\textbf{Platform and scene selection.}
We hosted the questionnaire on Zoho Forms and recruited
participants internationally through Prolific's AI Taskers
pool. The study included seven scenes from web videos:
bird singing, horse riding, karate, tap dancing, an
airplane, a sports car, and a running humanoid robot.
Two additional scenes from \ourdataset featured two
cars traveling in opposite directions and two people
clapping, respectively.

\noindent\textbf{Comparison design.}
For each of the nine scenes, we compared our method
against Sonic4D~\citep{xie2025sonic4d} and OmniAudio~\citep{liu2025omniaudiogeneratingspatialaudio}, yielding 18 baseline
comparisons. We additionally compared our full method
against the V2A and T2A variants on three scenes:
the airplane, two people clapping, and tap dancing.
This yielded six ablation comparisons and 24 video
pairs in total.

\noindent\textbf{Questionnaire.}
Each pair was accompanied by a one-sentence sound
source description specifying the main objects
and activities the users are expected to hear. Methods were presented as
\emph{Method 1} and \emph{Method 2}, without revealing
their identities. For every pair, participants answered
the required question:
\emph{``Which method, A or B, do you prefer overall
in terms of spatial audio?''}
For comparisons against Sonic4D, they additionally
answered:
\emph{``Which method, A or B, has better overall
visual quality?''}
This second question was included only for Sonic4D
because it uses its own visual generation backbone;
OmniAudio and our ablated variants share our
novel-view visual renderings.
Each question required selecting one of the two
methods, yielding 24 audio judgments and nine visual
judgments per participant.

\section{Implementation Details}
\label{app:implementation_details}

\subsection{Component-Wise Implementation}
\label{app:implementation_components}

\noindent\textbf{Video preprocessing and scene understanding.}
We use VGGT-$\Omega$~\citep{wang2026vggtomega} to estimate per-frame depth,
camera intrinsics, and camera poses, and SAM3~\citep{carion2026sam3segmentconcepts} to
segment and track foreground objects.
For the tracked FoV videos in \ourdataset, we retain
the known virtual-camera intrinsics and rotations
used during panorama-to-perspective projection,
together with the predicted depth.
We query Kimi K3~\citep{team2026kimi} with uniformly sampled video frames
to obtain structured scene descriptions, sound-source
attributes, and generation and separation prompts.
The VLM prompt is summarized in Fig.~\ref{fig:vlm-prompt}.

\noindent\textbf{Visual scene generation.}
For background generation, we aggregate the observed
background geometry into a partial panorama and use
GeoCalib~\citep{geocalib} for gravity alignment.
We complete missing regions using FLUX.2 Klein 9B~\citep{flux2}
with a panorama-outpainting LoRA~\citep{nomadoor2026flux2panorama}.
For \ourdataset, we instead use the captured full
background panoramas.
We use HYWorld 2.0~\citep{hyworld2026} for panorama-to-3D generation in
quantitative evaluations and Marble~\citep{worldlabs_marble} for qualitative
comparisons and demos.

Foreground generation follows the motion-type-specific
instantiations in Appendix~\ref{app:motion_instantiations},
with reconstruction and motion estimation details
provided in Appendix~\ref{app:foreground_implementation}.

\noindent\textbf{Foreground-background alignment.}
We follow the alignment procedure described in
Appendix~\ref{app:fg_bg_alignment}.
For scale estimation, we reject outliers in the
log-depth ratios of reliable background correspondences
before confidence-weighted fitting.
For outdoor scenes, scale estimation uses the lower
half of the reference panorama.
Local displacement estimation uses visible,
depth-consistent correspondences and rejects outliers
before averaging.

\noindent\textbf{Audio generation and grounding.}
We instantiate both the camera-space video-to-audio
branch and the object-space text-to-audio branches
with HunyuanVideo-Foley~\citep{shan2025hunyuanvideofoleymultimodaldiffusionrepresentation}, using its XXL configuration.
The camera-space branch receives the input video and the full-scene
audio description. By default, each object-space branch is conditioned only on its source-specific text prompt. We additionally enable video conditioning only when the VLM scene
analysis indicates that the source remains at an approximately
constant distance from the input camera throughout the video.
We use SAM-Audio-small~\citep{samaudio} for query-conditioned source
separation during camera-to-object synchronization.
Each static background sound is generated through
text-to-audio conditioning.
For non-global static sources, SAM3 segments the
corresponding regions in the background panorama,
which are associated with reconstructed background
geometry to obtain fixed spatial anchors.
Dynamic sources are associated with their reconstructed
object trajectories, and global ambience is retained
as a separate background track. 

Both generation branches use 100 Euler steps with a step size of 0.01 and classifier-free guidance of 6.0. Camera-space fusion uses $\eta_{\mathrm{cam}}=0.8$, while the object-space fusion uses $\eta_{\mathrm{obj}}=0.5$. For numerical stability, the denominator in the synchronized velocity update is clamped to $\max(1-\tau,0.05)$. Each source-separation call uses 32 midpoint steps with a step size of $1/32$, without classifier-free guidance. Separation results are recomputed at each synchronization step,
with hint caching disabled. Generated audio is decoded at 48\,kHz.

\subsection{Foreground Reconstruction and Motion Estimation}
\label{app:foreground_implementation}

\noindent\textbf{Rigid-body keyframe selection.}
We select up to 10 reliable keyframes per object for
MV-SAM3D reconstruction and FoundationPose registration.
We exclude frames with empty or insufficiently large
foreground masks, substantial occlusion, severe
image-boundary truncation, or blurred foreground observations.
Among the valid frames, we choose the one with the
largest foreground mask as the reconstruction reference.
The remaining keyframes are selected by farthest-point
sampling based on object positions and frame timestamps
to cover both spatial motion and time.

\noindent\textbf{Pose registration and trajectory completion.}
We register object poses at anchor frames using
FoundationPose and track bidirectionally between
consecutive anchors, averaging valid estimates from
both directions.
Before the first anchor and after the last anchor,
we track backward and forward, respectively.
Tracking stops when observations become unreliable
due to occlusion, insufficient object size, or
departure from the image.
For rigid-body objects, we fill gaps between reliable
poses by interpolating translations and rotations,
and complete trajectories outside this range using
constant-velocity translation extrapolation and
the nearest reliable rotation.
Available foreground masks provide additional position
and depth cues, and we refine the completed trajectories
using silhouette-containment and temporal constraints.

\noindent\textbf{Human representation.}
We store posed human Gaussians relative to the SMPL-X
pelvis and retain the pelvis trajectory separately.
This allows foreground-background alignment to adjust
global placement without modifying the recovered
articulation.

\subsection{VLM Prompt}
\label{app:vlm_prompt}

We jointly provide sampled video frames to the VLM
to obtain structured scene and sound-source descriptions.
The JSON response includes a static environment description,
an indoor/outdoor label, a full-scene audio description,
and source-specific attributes and prompts.
Each sound source is assigned a visual grounding label,
a target peak level, and a spatial type.
Dynamic foreground sources additionally receive
rigidity and support-surface labels, together with
identifiers linking their generation and separation
prompts to the corresponding reconstructed objects.
The support-surface labels also guide the layered
rendering described in Appendix~\ref{app:layered_rendering}.
Fig.~\ref{fig:vlm-prompt} summarizes the prompt instructions.

\begin{figure*}[tp]
  \centering
  \begin{minipage}{\textwidth}
  \begin{scriptsize}
\begin{verbatim}
TASK
You are given MULTIPLE images sampled from one VIDEO.
Infer plausible sounds from visible content and strongly implied environmental context.

OUTPUT
Return ONE JSON object with the following fields. Output JSON ONLY.
- "scene_static_prompt": one short sentence describing only the static environment.
  Exclude moving subjects, actions, and sounds.
- "environment": "indoor" or "outdoor".
- "video_description": 2-5 sentences describing the full plausible soundtrack,
  including foreground events, temporal activity, spatial behavior, and ambience.
- "sounds": 2-12 sound-source entries for grounding, loudness, and spatial type.
- "sync_sounds": 1-6 entries with generation prompts and separation queries.

FOR EACH ENTRY IN "sounds"
- "diffusion_prompt": <=10 simple words describing one sound source.
- "grounding_label": a 1-2 word visible object or host category.
  Ground implied or off-screen sounds on a visible host; mark them as non-foreground.
- "peak_db": integer negative dB for the target peak level; never greater than -6.
- "source_type": "point" for a localized source, "area" for an extended source,
  or "background" for global or diffuse ambience.
- "foreground": true only for a visible, principal dynamic foreground actor.

ADDITIONAL FIELDS FOR FOREGROUND ENTRIES ONLY
- "rigid": true for mostly rigid objects; false for deformable actors.
- "support_surface": a short visual label for the surface actually touched,
  e.g., "asphalt road", "wooden stage floor", or "treadmill belt".
  Name the topmost supporting surface. Use "none" for unsupported objects.
- "dynamic_source_id": a zero-based foreground instance identifier.
- "geometry_object_id": the corresponding reconstructed geometry identifier.

INSTANCE AND ORDERING RULES
- Create separate foreground entries for visually distinguishable instances.
- Order foreground entries first by perceived loudness, then environmental sources.
- Assign foreground identifiers consecutively in this order.
- Place exactly ONE global ambience entry last; avoid duplicate background beds.
- Static and global-background entries must omit foreground-only fields.

GLOBAL BACKGROUND BED
- "grounding_label": "global".
- "source_type": "background"; "foreground": false.
- Describe a realistic ambient bed, or silence if appropriate.
- Use a soft target peak level, typically -26 to -32 dB; use -120 for near silence.

FOR EACH ENTRY IN "sync_sounds"
- "diffusion_prompt": 1-4 sentences describing the source's timbre, rhythm,
  intensity, material interactions, and motion when relevant.
  For moving foreground sources, describe synchronization with visible activity.
- "seperation_prompt": a concise, source-focused separation query of 2-8 words.
  Preserve this exact field spelling.
- "motion_effect": boolean value indicating whether a Doppler effect should be observed.
- Foreground entries must repeat the matching "dynamic_source_id" and
  "geometry_object_id" from "sounds"; static/background entries omit these fields.
- Include major foreground events and important environmental sounds.
  Each entry must correspond to sources already listed in "sounds".

SELECTION AND DESCRIPTION RULES
- Infer realistic audio, rather than listing every visible object.
- Use evidence across frames and visible material or contact cues.
- Do not invent speech, music, horns, sirens, or animal sounds without support.
- Describe foreground motion and temporal activity when visually implied.
- Describe background sources as stable ambience where appropriate.
- Keep grounding labels concise and generation prompts audio-oriented.
- Do not mention camera movement in short source labels.
- Return valid JSON only, without markdown, comments, or additional text.
\end{verbatim}
  \end{scriptsize}
  \end{minipage}
  \caption{\textbf{VLM prompt for video-based scene understanding.}
  Condensed presentation of the instructions used to obtain
  scene descriptions, sound-source attributes, and generation
  and separation prompts. Examples and repeated validation
  instructions are omitted for brevity.}
  \label{fig:vlm-prompt}
\end{figure*}

\subsection{Interactive Viewer}
\label{app:interactive_viewer}

We implement the browser-based viewer using
Three.js\footnote{\url{https://threejs.org/}}
and the Spark Gaussian Splatting
renderer\footnote{\url{https://sparkjs.dev/}}.
The viewer loads background Gaussian layers, canonical
foreground Gaussians, object transformations, and
per-frame deformation attributes.
Rigid objects reuse their canonical Gaussians with
time-dependent global transformations.
For deformable objects, we additionally update Gaussian
positions and, where available, rotations and scales
from the exported deformation sequences.
All objects share a playback timeline, which also
controls audio playback and seeking.
Users can follow recorded camera paths or navigate
freely through the scene.

Interactive audio playback uses the Web Audio
API\footnote{\url{https://developer.mozilla.org/en-US/docs/Web/API/Web_Audio_API}}. The listener's position and orientation follow the
viewing camera. Dynamic source positions follow their
reconstructed trajectories, while static sources remain
at their grounded background anchors. Each localized
source is spatialized using HRTF-based panning with
distance-dependent attenuation, updating its spatial
cues as the source or listener moves. The spatialized
source contributions are mixed with global ambience
to produce binaural output.

\subsection{Layered Rendering}
\label{app:layered_rendering}

We use layered rendering to address two types of
incorrect foreground occlusion.
First, inaccuracies in object poses and reconstructed
supporting surfaces can leave parts of an object below
the surface, making feet or wheels appear buried even
after foreground-background alignment.
Second, sky Gaussians reconstructed by Marble may lie
closer to the viewer than distant foreground objects,
causing the sky to incorrectly occlude them.
We therefore separate the background into sky,
supporting surfaces, and other scene geometry, and
introduce a depth tolerance when compositing foreground
objects against the sky and supporting surfaces.
Other scene geometry, such as trees and walls, retains
depth-based occlusion.

\noindent\textbf{Background layer construction.}
After alignment, we partition the background Gaussians
into sky, supporting surfaces, and remaining background
geometry.
For outdoor scenes, we segment the sky in the background
panorama using SAM3 and transfer its mask to Gaussians
by projecting their centers into the panorama.
Indoor scenes omit the sky layer.

To identify supporting surfaces, we use the VLM-provided
surface labels as SAM3 prompts and combine these semantic
masks with contact geometry.
We collect the lowest foreground Gaussian points along
the aligned trajectories and identify nearby background
Gaussians at compatible heights as contact seeds.
The support region grows from these seeds along a
nearest-neighbor graph subject to spatial and height
continuity constraints.
When the semantic mask agrees with the contact seeds,
it restricts the region to the named surface near the
trajectory.
Objects labeled as unsupported, such as a plane
flying through the sky, do not contribute contact seeds.
All background Gaussians outside the sky and support
regions form the remaining geometry layer, which retains
potential occluders such as trees, pillars, and walls.

\noindent\textbf{Depth-aware compositing.}
The viewer renders three groups separately:
foreground objects, sky and supporting surfaces, and
the remaining background geometry.
For each group, we render both color and an
alpha-weighted view-depth image using the same splat
ordering and blending.
The groups are then composited per pixel.

Let $d_{\mathrm{fg}}$, $d_{\mathrm{sup}}$, and
$d_{\mathrm{occ}}$ denote the rendered depths of the
foreground, the combined support/sky group, and the
remaining background geometry, respectively.
Foreground visibility relative to the remaining
background follows the usual depth comparison,
$d_{\mathrm{fg}} < d_{\mathrm{occ}}$.
Against the support/sky group, we introduce a
depth tolerance $\epsilon = 2D$ and favor the foreground
when $d_{\mathrm{fg}} < d_{\mathrm{sup}} + \epsilon$,
where $D$ is the foreground object's bounding-box
diagonal length after applying its global scale.
We soften this transition using a smooth blending
weight around the threshold.
This reduces the disappearance of foreground contact
regions into the supporting surface while retaining
depth-based occlusion by other scene geometry.
The tolerance is expressed in scene units relative
to foreground size.

Layered compositing complements the geometric alignment
by correcting residual visibility artifacts at display
time. It leaves the reconstructed Gaussian positions,
object trajectories, and acoustic geometry unchanged.

\end{document}